%% file: acl_latex.tex
\pdfoutput=1

\documentclass[11pt]{article}

\usepackage[]{acl}

\usepackage{times}
\usepackage{latexsym}
\usepackage[T1]{fontenc}
\usepackage[utf8]{inputenc}
\usepackage{microtype}

\usepackage{inconsolata}
\usepackage{graphicx}

\usepackage{xcolor,soul}
\usepackage{booktabs}
\usepackage{enumitem}
\usepackage{etoolbox}
\usepackage{colortbl}
\usepackage{transparent}
\usepackage{amssymb}
\usepackage{dsfont}
\usepackage{amsmath}
\usepackage{url}
\usepackage{adjustbox}
\usepackage{listings}
\usepackage{float}
\usepackage{array}
\usepackage{authblk}

\newcolumntype{|}{!{\color{white}\vline width 10pt}}
\newcolumntype{/}{!{\color{white}\vline width 5pt}}

\title{Summary of a Haystack:\\A Challenge to Long-Context LLMs and RAG Systems}

\newcommand{\benchmark}{SummHay}

\author{
  \quad \textbf{Philippe Laban}\Thanks{~~Equal contribution}
  \quad \textbf{Alexander R. Fabbri}\textsuperscript{$*$}
  \quad \textbf{Caiming Xiong}
  \quad \textbf{Chien-Sheng Wu} \\
  Salesforce AI Research \\
  \{plaban, afabbri, cxiong, wu.jason\}@salesforce.com \\
}

\begin{document}
\maketitle

\input{0-abstract}
\input{1-introduction}
\input{2-related-work}
\input{3-framework}
\input{4-evaluation}
\input{5-experiments}
\input{6-ablations}
\input{7-discussion}
\input{8-conclusion}

\input{9-limitations}

\bibliography{anthology,custom}

\appendix
\input{10-appendix}
\end{document}

%% file: 0-abstract.tex
\begin{abstract}
LLMs and RAG systems are now capable of handling millions of input tokens or more. 
However, evaluating the output quality of such systems on long-context tasks remains challenging, as tasks like Needle-in-a-Haystack lack complexity.
In this work, we argue that summarization can play a central role in such evaluation. We design a procedure to synthesize Haystacks of documents, ensuring that specific \textit{insights} repeat across documents.
The ``Summary of a Haystack'' (SummHay) task then requires a system to process the Haystack and generate, given a query, a summary that identifies the relevant insights and precisely cites the source documents.
Since we have precise knowledge of what insights should appear in a haystack summary and what documents should be cited, we implement a highly reproducible automatic evaluation that can score summaries on two aspects -- Coverage and Citation.
We generate Haystacks in two domains (conversation, news), and perform a large-scale evaluation of 10 LLMs and corresponding 50 RAG systems.
Our findings indicate that SummHay is an open challenge for current systems, as even systems provided with an Oracle signal of document relevance lag our estimate of human performance (56\%) by 10+ points on a Joint Score.
Without a retriever, long-context LLMs like GPT-4o and Claude 3 Opus score below 20\% on SummHay.
We show SummHay can also be used to study enterprise RAG systems and position bias in long-context models.
We hope future systems can equal and surpass human performance on SummHay.

\end{abstract}

%% file: 1-introduction.tex
\section{Introduction}


Recent progress in efficient attention mechanisms has led to the expansion of the context length of large language models \cite{beltagy2020longformer,su2024roformer}.
Previous state-of-the-art models such as T5 \cite{raffel2020exploring} and BART \cite{lewis2019bart} were limited to input contexts of 512 or 1024 tokens, while the latest models such as Claude-3 or Gemini-1.5-pro \cite{reid2024gemini} can process sequences of hundreds of thousands or millions of tokens.

Another paradigm, Retrieval Augmented Generation (RAG) \cite{lewis2020retrieval,guu2020retrieval}, has emerged as an alternative to these long-context LLMs, proposing a pipelined approach in which a retriever dynamically selects the context relevant to a given input query, alleviating the need for the generator to process long contexts directly.

Although both RAG and long-context LLMs offer to solve the common problem of answering queries over a large corpus of text, a direct comparison on a common task is still lacking, and evaluating such systems is an open challenge.
%
Some recent work has popularized tests such as the Needle-in-a-Haystack task \cite{kamradt2023}, which requires models to identify a small piece of information in a large document. However, these tasks do not offer the complexity needed to differentiate the capabilities of the latest generation of large-language models, with several state-of-the-art models achieving near-perfect performance.

In this work, we propose to leverage the task of summarization as a testbed for evaluating long-context models and RAG systems. Summarization requires reasoning over a long context and a careful understanding of the relative importance of content.
However, most prior work on summarization evaluation, particularly in evaluating the relevance of summaries, has focused on single-document summarization or tasks in which the input content is on the order of 1,000-2,000 tokens \cite{laban2020summary,fabbri2021summeval,Bhandari-2020-reevaluating,liu2022revisiting}. 
Apart from \citet{chang2023booookscore}, which focuses on summary coherence for 100k-token books, other evaluation work on longer conversational and multi-document news summarization is still often limited to around 10k tokens \cite{zhong-etal-2021-qmsum,huang2023embrace}.

A central problem is that summarization evaluation often relies on low-quality reference summaries and automatic metrics that do not correlate well with human judgments.
Within reference-based evaluation, a candidate summary is compared to a gold-standard reference summary, with the optics that a higher overlap between the candidate and reference summary indicates higher quality.
This paradigm may limit evaluation reliability, due to the lack of gold-standard references, particularly in long-context settings where obtaining high-quality summaries would be prohibitively expensive.
Furthermore, automatic metrics may still fail to correlate well with human judgments with respect to these references; despite the human-validated pipeline of \citet{huang2023embrace}, the best automatic metric for content coverage in that study has a correlation of just 0.37 with human judgment.

In this work, we address these limitations through synthetic data generation.
An overview of our Framework is found in Figure \ref{fig:haystack_synthesis}.
We propose data synthesis programs to generate a large corpus of documents (the ``Haystack'') on a given topic.
By enforcing that certain units of information (``insights''), categorized according to various subtopics, repeat within Haystack documents, and precisely controlling which insights occur in which documents, we can automatically derive the relevant insights within the Haystack for a given search query.
A system completing the Summary of a Haystack (SummHay) task must then summarize insights relevant to a search query and cite the source documents of each insight.
These summaries can be evaluated based on whether they cover the expected reference insights, and cite precisely and thoroughly the source documents.

Our first contribution is a procedure for generating Haystacks in two domains: conversations and news articles. Section~\ref{sec:framework} details the carefully-designed pipeline to ensure the feasibility and validity of the task.
A Haystack typically contains 100 documents on a topic, totaling approximately 100k tokens. 
We generate a total of 10 Haystacks, each coupled with roughly 10 queries, for a total of 92 SummHay tasks. Our pipeline can be scaled and applied to other domains.

Our second contribution develops SummHay's evaluation protocol, centering on evaluating system outputs on their Coverage of reference insights, and the quality of their Citation. A manual annotation confirms strong reproducibility of the protocol among knowledgeable annotators (0.77 correlation). We then experiment with LLM-based evaluation, finding that although the level of correlation is slightly lower (0.71), evaluation cost is reduced by a factor of almost 50.

Our third contribution is an estimate of human performance on SummHay and a large-scale evaluation of 50 RAG systems and 10 long-context LLMs.
Our findings indicate that: (1) SummHay is challenging for all systems we evaluate, with all models significantly below our estimate of human performance, even when given oracle signals of document relevance; (2) non-trivial trade-offs exist when choosing between a RAG pipeline and a long-context LLM, with RAG systems typically improving citation quality, at the cost of insight coverage, (3) using advanced RAG components (e.g., Cohere's Rerank3) leads to end-to-end performance boosts on the task, confirming that SummHay is a viable option for holistic RAG evaluation, (4) a positional bias experiment on SummHay confirms the \textit{lost in the middle} phenomenon, demonstrating that most LLMs are biased towards information at the top or bottom of the context window.

We open-source our dataset and evaluation methodology\footnote{\url{https://github.com/salesforce/summary-of-a-haystack}}.
A system that achieves a high score on SummHay can reliably reason over large corpora of documents, detect and summarize insights, and accurately cite its sources. We anticipate that although our findings indicate that human performance is still out of reach, future systems can achieve and surpass such performance, providing more reliable and trustworthy answer engines.

%% file: 2-related-work.tex
\section{Related Work}

\begin{figure*}[ht]
    \centering
    \includegraphics[width=0.98\textwidth]{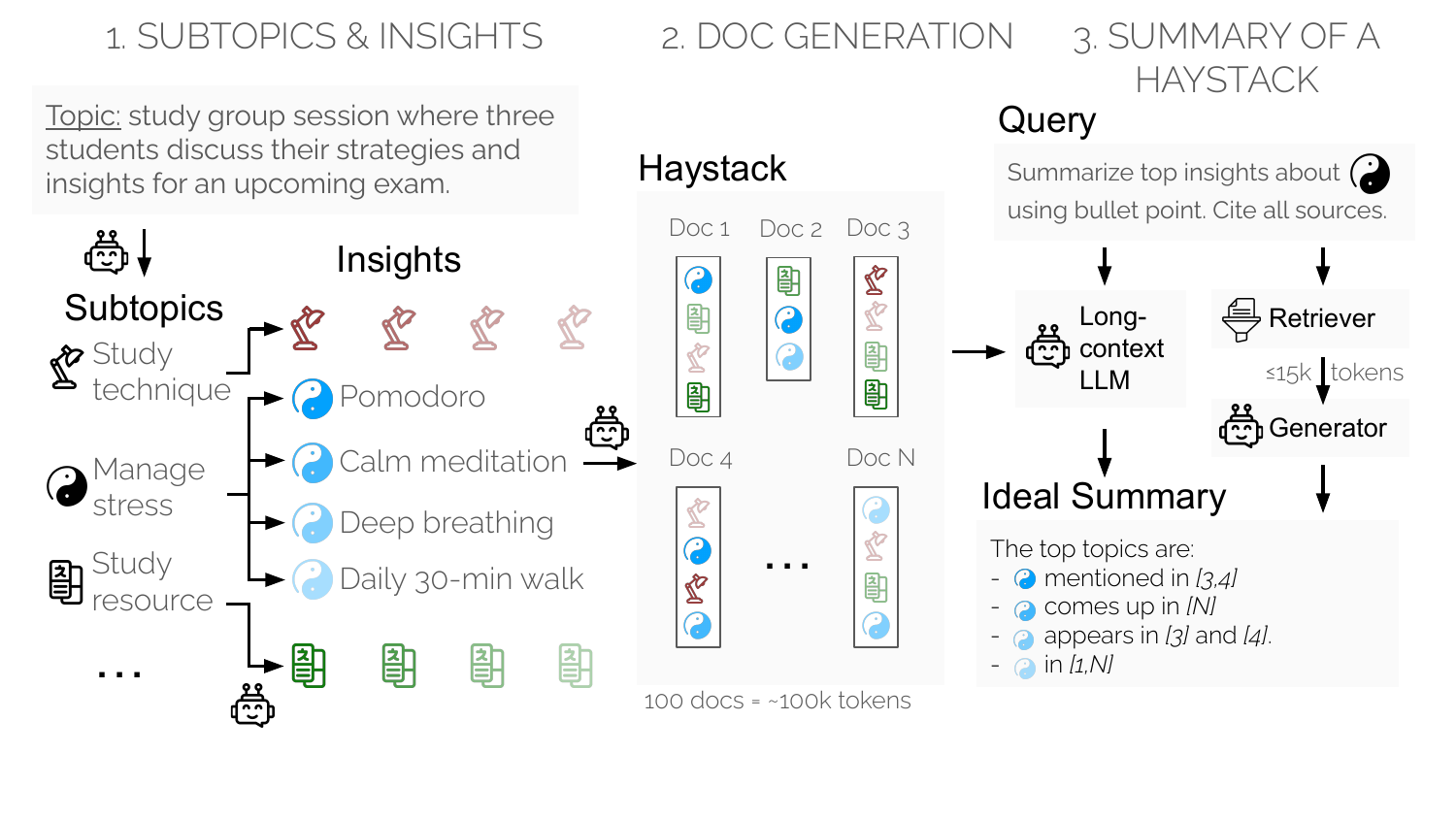}
    \vspace{-0.5in}
    \caption{Diagram illustrating the steps to synthesize a Haystack of documents given an input scenario: subtopic and insight creation followed by document generation. Once a Haystack is synthesized, it can be used to benchmark LLMs / RAG systems on query-focused summarization tasks.}
    \label{fig:haystack_synthesis}
\end{figure*}

\subsection{Summarization Evaluation}
Existing work in summarization relevance, or coverage, evaluation has largely focused on the short-input, single-document setting \cite{gao-wan-2022-dialsummeval,fabbri-etal-2021-summeval}.
Extending to long input evaluation, recent work has performed meta-evaluation on coherence in book summarization \cite{chang2023booookscore} and faithfulness across several domains \cite{krishna2023longeval,min2023factscore,zhang2024fine}.
For coverage evaluation, recent work has studied content selection for book summarization \cite{kim2024fables}, evaluated a two-step extract-evaluate framework \cite{wu2023less}, and compared the correlation of LLM metrics in  coverage \cite{huang2023embrace}.
We leverage summarization relevance as a test-bed for long-context evaluation, and we focus on our synthetic creation pipeline and the simplified relevance evaluation that results in a high-correlation automatic metric.
%

%
\subsection{Long-Context LLM Evaluation}
Needle-in-a-haystack \cite{kamradt2023} was proposed to assess the long-context recall ability of LLMs. 
Subsequent work has analyzed the effect of needle placement \cite{machlab2024llm} and multi-needle \cite{multineedle} and multi-modal variations \cite{song2024milebench,reid2024gemini}.
\par 
Additionally, several long-context evaluation benchmarks have been created, for example by building upon and revising existing tasks and datasets \cite{bai2023longbench,an2023eval}
Some work proposes ways to extend the context length of shorter-context datasets; \cite{kuratov2024search,kwan2023m4le}, while other work addresses data contamination in long-context settings \cite{ni2024xl,dong2023bamboo}.
%
Several papers introduce synthetic data in additional to existing tasks \cite{shaham2023zeroscrolls,zhang2024infty}, which can prove to be more difficult for current models, as seen in \citet{bai2023longbench}.
Our benchmark focuses on synthetic data on the scale of 100k input tokens, and as opposed to existing synthetic tasks centered largely around retrieving, counting, or sorting, our summarization task requires aggregating and non-trivial reasoning over the long-context.

\subsection{Attribution Evaluation}
Several benchmarks have emerged to study the ability of LLMs to ground their generation with citations \cite{li2023survey}.
AttributionBench \cite{li2024attributionbench} aggregates 7 existing attribution datasets, including Hagrid \cite{kamalloo2023hagrid}, consisting of generative answers to questions annotated by humans for attribution, and AttrEval-GenSearch \cite{yue2023automatic}, which categorizes attribution into three levels of support.
Attribution evaluation has also been performed along sources beyond documents such as knowledge graphs \cite{hu2024benchmarking,li2023towards} and for tasks such as long-form question-answering \cite{chen2023understanding}. 
Specific to summarization, Seahorse \cite{clark2023seahorse} collects annotations for summary attribution in the short-context setting. 
In this paper we study attribution, or citation, in the context of long-input summarization. 
Due to our synthetically generated data, we can trace reference insights to their sources and directly evaluate summary citations. 


%% file: 3-framework.tex
\section{Summary in a Haystack Framework} \label{sec:framework}

\begin{figure*}[ht]
    \centering
    \includegraphics[width=0.95\textwidth]{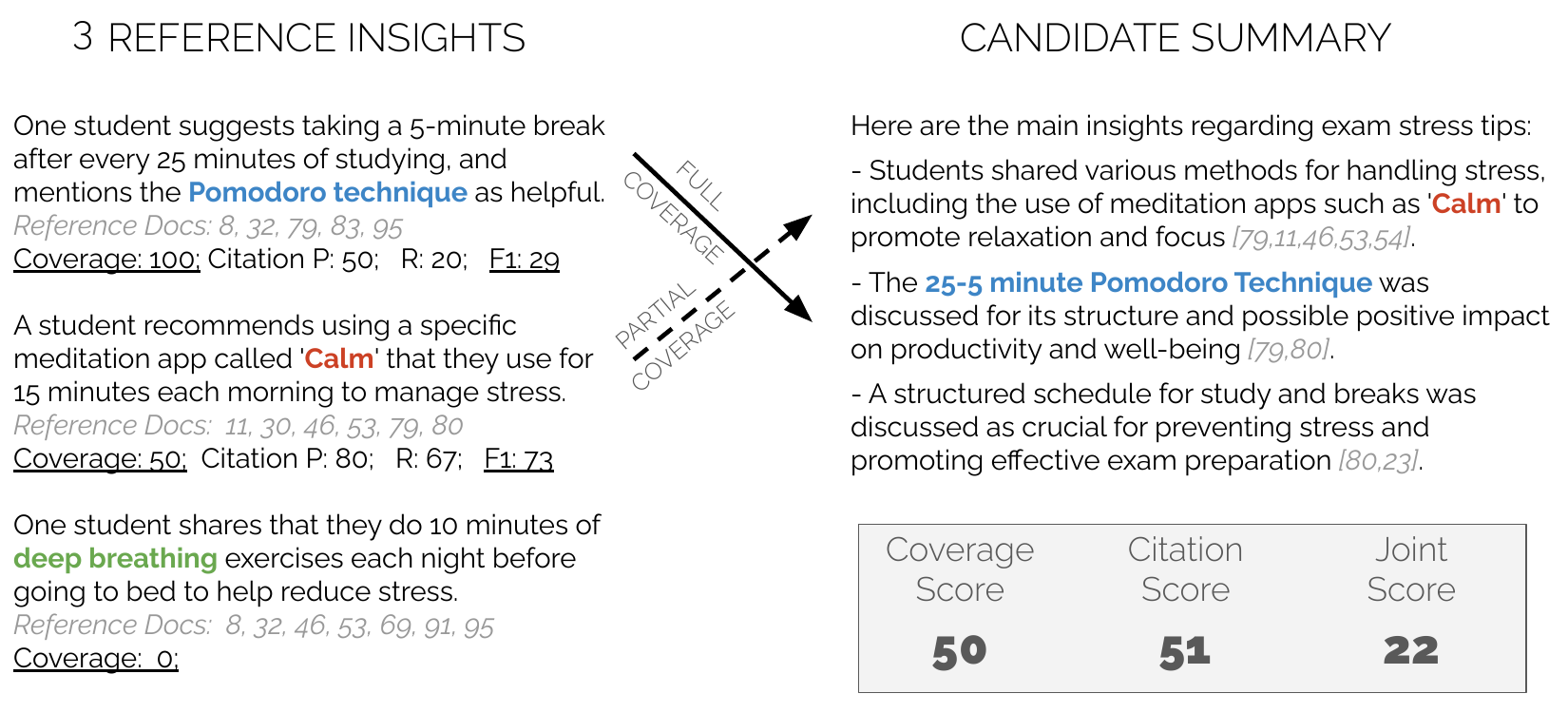}
    \caption{Example evaluation of a candidate summary (right) for its coverage of reference insights (left). Each reference insight is assigned a \texttt{Coverage Score} by mapping it to a single candidate bullet. A mapped bullet's citations are used to calculate the \texttt{Citation Score}. The total score is the average across reference insights. See Appendix~\ref{app:additional_examples} for four additional examples.}
    \label{fig:eval_figure}
\end{figure*}

Figure~\ref{fig:haystack_synthesis} illustrates the process of synthesizing Haystack data, and the task, which we now detail.

\subsection{Preliminaries}

In SummHay, as in the needle-in-a-haystack task, the LLM responds to a query, but here it must generate a long-form answer (200-300 words) that requires identifying and summarizing insights that repeat across documents and citing source documents. The task resembles long-form question-answering \cite{fan-etal-2019-eli5} and query-focused summarization \cite{zhong-etal-2021-qmsum,vig-etal-2022-exploring}.

In the following section, we describe the Haystack synthesis, the steps taken to ensure the quality of our benchmark, and the task framing.

\subsection{Haystack Generation}
\label{subsec:haystack_gen}
\par \textbf{Subtopic and Insight Generation (Figure~\ref{fig:haystack_synthesis}, left)} 
One of the main motivations behind synthetically generating documents is to precisely control information distribution in the documents.

A Haystack centers around a \textit{topic} (e.g., ``three students discuss strategies for an upcoming exam''). The first step generates a list of potential subtopics that can occur in documents about the topic. Subtopics are generic (e.g., students discussing study techniques, or how to manage stress). There are two subtopic requirements: (1) each subtopic should be \textit{distinctive} and unique, such that no two subtopics overlap thematically, and (2) subtopics should be \textit{expandable} into at least three distinct insights that are specific to the subtopic. Appendix~\ref{app:haystack_verification} goes over the quality assurance to ensure the satisfaction of these requirements.

%


In a second step, each subtopic gets instantiated into a list of specific \textit{insights}, or facts that will be placed into the documents of the Haystack. Insights are defined as statements that contain specific information that may appear in a document about a given subtopic.
Insights are expected to mention a number, a date, or an entity.
For example, in the ``Managing stress'' subtopic, an insight can be: ``a student explaining what the 25-5 Pomodoro technique is to the others''.
Crucially, insights should be specific, independent of each other, and solely relevant to a single subtopic. Appendix~\ref{app:insight_verification} goes over the quality assurance to ensure insight quality.

The idea of breaking down documents into smaller information units has proven beneficial in recent work for both automatic and human evaluation \cite{min2023factscore, liu2022revisiting}.
Concretely, we use an LLM to generate subtopics and insights, optionally include context documents to provide seed ideas to the LLM, and aim to generate 10 subtopics, each with about 5-10 insights.



\par \textbf{Document Generation (Figure~\ref{fig:haystack_synthesis}, center)}

The Haystack is synthesized one document at a time.
For each document, we randomly select a set of insights across subtopics, and instruct an LLM to generate a document that must include all selected insights.
The number of insights to include per document varies based on the domain, targeting 750 words of content per document (or roughly 1,000 tokens). By generating 100 documents, a Haystack totals on the order of 100k tokens. Appendix~\ref{app:haystack_verification} details quality assurance that ensures documents are realistic and unique, and that each insight occurs within 5+ documents.


Evaluation of the SummHay task relies on precise knowledge of the mapping between insights and documents in the Haystack. We implement over five domain-specific verification processes during the synthesis of the Haystack to ensure that the expected mapping is sound.
Manual inspection and the high performance of human annotators on the final task, shown in Section~\ref{subsec:human_perf} provide evidence of the quality of the resulting Haystacks.



\subsection{Haystack Summarization (Figure~\ref{fig:haystack_synthesis}, right)}
Having generated Haystacks following the above protocol, we can now leverage them for the Summary of a Haystack task.
Using an LLM, we transform each subtopic into a query (e.g., ``What do the students discuss regarding stress management?''). 

Each system completing the task is instructed to generate a summary to answer the query (which focuses on a single subtopic), which must be in bullet-point format. Crucially, we instruct the LLM on the number of bullet points that the summary should contain, which matches the number of insights of the subtopic.
Although this can appear as a simplifying assumption, this important design choice allows us to control for the length of generated summaries, which are a known confounding factor in summarization evaluation \cite{liu2022revisiting}.
We find in Section~\ref{sec:results} that this choice effectively controls the length of generated summaries. The prompt also instructs the system to cite source documents in each of its bullet points, in a specific bracketed format (e.g., [1,2]), using document identifiers provided in the Haystack.
The bullet-point structure and specific citation format are the foundation for our evaluation, detailed in Section~\ref{sec:evaluation}.

\subsection{\benchmark\ Benchmark}
\label{subsec:benchmark}
We instantiate the above protocol across two domains: conversations and news, as these two domains are common test beds for summarization \cite{DBLP:conf/nips/HermannKGEKSB15,gliwa-etal-2019-samsum}. 
For each domain, we generate 5 Haystacks, and the average Haystack length is 93k tokens.
Each Haystack consists of on average 9.20 subtopics, each averaging 6.75 insights, for a total of 62 insights per topic.
For the news domain, we leverage the documents from \citet{huang2023embrace} as the seed context documents.
Regarding LLM choice, we rely on a combination of GPT-3.5 and GPT-4o and specify additional details in Appendix \ref{app:haystack_verification}.

%% file: 4-evaluation.tex
\input{tables/auto_eval_results}

\section{Evaluation Protocol} \label{sec:evaluation}

We first define task metrics (illustrated in Figure~\ref{fig:eval_figure}), establish the reproducibility of manual annotation, and then assess the quality of automated evaluation.

\subsection{Evaluation Metrics}

\par \textbf{Coverage Metric}
Given a candidate subtopic summary, we extract each bullet point (split on line breaks) and want to measure the overlap between these bullets and the subtopic's reference insights. To do so, we iterate over each reference insight and assess whether the reference insight is \textit{fully}, \textit{partially}, or \textit{not} covered in any of the candidate bullet points. For each insight, the summary receives a score of 100 for full coverage, 50 for partial coverage, and 0 otherwise. The final coverage score of a summary is the average coverage on all the insights of the subtopic, such that it ranges from 0 to 100. In Figure~\ref{fig:eval_figure}, the top reference insight is fully covered by the second candidate bullet, the second insight is partially covered by the first candidate bullet, and the third insight is not covered in the summary. The Coverage Score is: $(100 + 50 + 0) / 3 = 50$.

%
%
\par \textbf{Citation Metric}
Because documents of the Haystack are synthesized, each reference insight can be traced to a gold-standard set of documents that contain the insight. When a summary's bullet covers a reference insight, we can compare generated citations to this reference set of cites.
For each partially or fully covered reference insight, cited documents are extracted from the paired summary bullet point using a regular expression (\texttt{[.*]}), and we measure the precision and recall between the generated and gold-standard cites.
The Citation Score of a reference insight is calculated as the F1 score of the precision and recall, giving equal weight to both. In short, a system must be both precise and thorough in its citing to achieve a high Citation Score.
The Citation Score of an entire summary is then the average insight Citation Score of all reference insights that were covered by the system.
In Figure 2, the average Citation F1 of the two covered bullets is: $(29 + 73) / 2 = 51$. 

%
\par \textbf{Joint Metric}
The Joint Metric pieces Coverage and Citation Scores together, measuring whether a candidate summary both covers the expected insights and cites documents appropriately. The Joint Score of a summary is calculated by iterating over each reference insight and multiplying its coverage score and citation scores (assigning a Citation Score of 0 in case the insight is not covered). The Joint Score of a summary ranges from 0 to 100. In Figure~\ref{fig:eval_figure}, the summary's Joint Score is: $(100*0.29 + 50*0.73 + 0*0) / 3 = 21.8$. Appendix~\ref{app:additional_examples} provides four additional examples on the same subtopic with added details.

\input{tables/main_results} 

\subsection{Annotation Reproducibility} \label{sec:annot_repro}

To establish the reproducibility of the evaluation protocol, two authors of the paper and two professional annotators independently annotated a common subset of 35 summaries, annotating for the coverage of 240 insights within the summaries. The Manual Annotation row of Table~\ref{tab:auto_eval_results} reports the inter-annotator agreement levels on the 240 coverage judgments. Coverage Score averaged a correlation of 0.77 across pairs of annotators, indicating a strong level of agreement between participants. When annotators agree that a reference insight is covered, they agree on which candidate bullet originates the coverage in 95\% of cases (\texttt{Linking Accuracy}). In short, \textbf{annotators have strong agreement on which reference insights are covered by a given candidate summary and also agree on the specific bullets in the candidate summary that cover each insight.}

Annotating a single summary takes 4 minutes on average. To reduce evaluation costs and scale experiments, we next investigate LLM-based evaluation as an alternative to annotation.

\subsection{Automatic Metric Validation}

We recruited two professional annotators to annotate 200 candidate summaries (100 for each of the news and conversational domains) paired with reference insights. Annotators were compensated at \$25/hour, and annotation required a total of 13 hours of work, for a total of \$325.

We prepared a prompt that contains task instructions and an example summary which has a fully covered, a partially covered, and an uncovered insight (see Appendix~\ref{app:eval_prompt}). We evaluated 5 LLMs as evaluators: GPT3.5, Claude 3 Haiku, Opus, GPT-4o, and Gemini-1.5-pro. In Table~\ref{tab:auto_eval_results}, we report evaluator performance in terms of correlation on the insight-level coverage scores, linking accuracy, which measures whether LLMs can attribute the coverage to the correct bullet point, and the cost of evaluating the 200 summaries. We find that two models, GPT-4o and Gemini-1.5-pro achieve a strong positive correlation (0.7+) with the human annotation. We select the GPT-4o model as our evaluator, as it achieves high evaluation correlation at a fraction of the cost of Gemini-1.5-pro, and does not have strict rate limits in place. We attempt one improvement by preparing a prompt with 9 few-shot examples (three of fully, partially, and uncovered insights each), and report the result as \texttt{GPT-4o (9FS)}. Although the increased number of examples does lead to minor correlation improvements, it comes at a large cost increase, and thus we finalize the auto-evaluation setting as using the original prompt and the GPT-4o evaluator.

In Appendix~\ref{app:auto_eval_bias}, we assess whether automatic evaluation using GPT-4o could cause two types of biases: first, whether it could systematically favor or disfavor summaries generated by a family of models (such as GPTs), and second whether it could be partial to summaries of a certain length. We find no sign of systematic bias in the LLM-based evaluation in the case of our protocol.

%% file: tables/auto_eval_results.tex
\begin{table}[]
    \centering
    \resizebox{0.44\textwidth}{!}{%
    \begin{tabular}{lcccc}
    
    \textbf{Method} & \textbf{Cov. Corr.} & \textbf{Link Acc.} & \textbf{Cost (\$)} \\
    \hline
    Manual Annot. & 0.770 & 95.0 & \$325 \hspace{0.05cm} \\
    \hline
    Gemini-1.5-pro & \textbf{0.751} & \textbf{89.3} & \$15.1 \\
    GPT-4o (9FS) & 0.719 & 89.2 & \$26.1 \\
    \underline{\textbf{GPT-4o}} & 0.716 & 88.9 & \$6.9 \hspace{0.1cm} \\
    Claude 3 Opus & 0.677 & 87.9 & \$23.8 \\
    Claude 3 Haiku & 0.498 & 87.7 & \$0.4 \hspace{0.1cm} \\
    GPT3.5 & 0.495 & 86.7 & \$1.3 \hspace{0.1cm} \\
    \bottomrule
    \end{tabular}
    }
    \caption{Reproducibility and cost of manual and automated evaluation for SummHay. We compute coverage correlation, linking accuracy, and evaluation cost.}
    \label{tab:auto_eval_results}
\end{table}

%% file: tables/main_results.tex
\begin{table*}[ht]
    \centering
    \renewcommand{\arraystretch}{1.3} 
    \resizebox{0.98\textwidth}{!}{%
    \begin{tabular}{r|cccccc/c|cccccc/c|cccccc/cc}

    & \multicolumn{7}{c}{\Large{\textbf{Coverage Score ($\uparrow$)}}} & \multicolumn{7}{c}{\Large{\textbf{Citation Score  ($\uparrow$)}}} & \multicolumn{7}{c}{\Large{\textbf{Joint Score ($\uparrow$)}}} \\

    \cmidrule(r{12pt}){2-8} \cmidrule(r{12pt}){9-15} \cmidrule(){16-22}
    \textbf{Summarizer} & Rand & Vect & LongE & KWs & RR3 & Orac & Full & Rand & Vect & LongE & KWs & RR3 & Orac & Full & Rand & Vect & LongE & Kws & RR3 & Orac & Full & \#$W_{b}$ \\
    \cmidrule(lr{15pt}){1-1} \cmidrule(r{5pt}){2-7} \cmidrule(lr{12pt}){8-8} \cmidrule(r{5pt}){9-14} \cmidrule(lr{15pt}){15-15} \cmidrule(r{5pt}){16-21} \cmidrule(l){22-22} \cmidrule(lr){23-23}

    GPT3.5 &\cellcolor[rgb]{0.97, 0.98, 1.00} 36.2 &\cellcolor[rgb]{0.89, 0.92, 0.99} 45.8 &\cellcolor[rgb]{0.88, 0.92, 0.99} 46.0 &\cellcolor[rgb]{0.86, 0.91, 0.98} 48.4 &\cellcolor[rgb]{0.83, 0.88, 0.98} 51.9 &\cellcolor[rgb]{0.79, 0.86, 0.98} 56.2 & -- &\cellcolor[rgb]{0.89, 0.69, 0.69} 9.3 &\cellcolor[rgb]{0.92, 0.76, 0.76} 15.2 &\cellcolor[rgb]{0.92, 0.76, 0.76} 15.0 &\cellcolor[rgb]{0.92, 0.77, 0.77} 15.9 &\cellcolor[rgb]{0.93, 0.78, 0.78} 16.8 &\cellcolor[rgb]{0.95, 0.87, 0.87} 23.0 & -- &\cellcolor[rgb]{0.87, 0.61, 0.61} 3.6 &\cellcolor[rgb]{0.88, 0.66, 0.66} 7.3 &\cellcolor[rgb]{0.88, 0.66, 0.66} 7.2 &\cellcolor[rgb]{0.89, 0.67, 0.67} 7.9 &\cellcolor[rgb]{0.89, 0.68, 0.68} 9.0 &\cellcolor[rgb]{0.91, 0.74, 0.74} 13.2 & -- & 28.2 \\
Claude 3 Haiku &\cellcolor[rgb]{0.85, 0.90, 0.98} 49.9 &\cellcolor[rgb]{0.71, 0.80, 0.97} 64.9 &\cellcolor[rgb]{0.73, 0.82, 0.97} 62.3 &\cellcolor[rgb]{0.72, 0.81, 0.97} 63.4 &\cellcolor[rgb]{0.70, 0.79, 0.97} 66.6 &\cellcolor[rgb]{0.66, 0.74, 0.93} 72.1 &\cellcolor[rgb]{0.73, 0.82, 0.97} 62.3 &\cellcolor[rgb]{0.91, 0.74, 0.74} 13.4 &\cellcolor[rgb]{0.96, 0.89, 0.89} 25.1 &\cellcolor[rgb]{0.96, 0.90, 0.90} 25.5 &\cellcolor[rgb]{0.97, 0.91, 0.91} 26.5 &\cellcolor[rgb]{0.98, 0.94, 0.94} 28.8 &\cellcolor[rgb]{0.98, 0.99, 1.00} 35.6 &\cellcolor[rgb]{0.91, 0.75, 0.75} 14.1 &\cellcolor[rgb]{0.88, 0.66, 0.66} 7.1 &\cellcolor[rgb]{0.93, 0.79, 0.79} 17.4 &\cellcolor[rgb]{0.93, 0.79, 0.79} 17.2 &\cellcolor[rgb]{0.93, 0.80, 0.80} 17.7 &\cellcolor[rgb]{0.94, 0.83, 0.83} 20.1 &\cellcolor[rgb]{0.97, 0.91, 0.91} 26.8 &\cellcolor[rgb]{0.89, 0.69, 0.69} 9.2 & 31.9 \\
GPT4-turbo &\cellcolor[rgb]{0.85, 0.90, 0.98} 49.4 &\cellcolor[rgb]{0.75, 0.83, 0.97} 61.0 &\cellcolor[rgb]{0.79, 0.85, 0.98} 56.7 &\cellcolor[rgb]{0.75, 0.82, 0.97} 61.2 &\cellcolor[rgb]{0.74, 0.82, 0.97} 61.8 &\cellcolor[rgb]{0.69, 0.79, 0.96} 67.1 &\cellcolor[rgb]{0.78, 0.85, 0.97} 57.9 &\cellcolor[rgb]{0.93, 0.80, 0.80} 17.9 &\cellcolor[rgb]{0.98, 0.94, 0.94} 28.6 &\cellcolor[rgb]{0.98, 0.93, 0.93} 28.1 &\cellcolor[rgb]{0.99, 0.97, 0.97} 31.1 &\cellcolor[rgb]{0.99, 0.98, 0.98} 31.8 &\cellcolor[rgb]{0.93, 0.95, 0.99} 41.4 &\cellcolor[rgb]{0.87, 0.64, 0.64} 5.5 &\cellcolor[rgb]{0.89, 0.69, 0.69} 9.6 &\cellcolor[rgb]{0.93, 0.81, 0.81} 18.7 &\cellcolor[rgb]{0.93, 0.79, 0.79} 16.9 &\cellcolor[rgb]{0.94, 0.83, 0.83} 20.1 &\cellcolor[rgb]{0.94, 0.83, 0.83} 20.6 &\cellcolor[rgb]{0.98, 0.94, 0.94} 28.9 &\cellcolor[rgb]{0.86, 0.61, 0.61} 3.2 & 37.9 \\
Command-r &\cellcolor[rgb]{0.87, 0.91, 0.99} 47.0 &\cellcolor[rgb]{0.80, 0.86, 0.98} 54.8 &\cellcolor[rgb]{0.82, 0.87, 0.98} 53.5 &\cellcolor[rgb]{0.79, 0.86, 0.98} 56.0 &\cellcolor[rgb]{0.80, 0.86, 0.98} 55.2 &\cellcolor[rgb]{0.75, 0.83, 0.97} 60.4 &\cellcolor[rgb]{0.84, 0.89, 0.98} 50.3 &\cellcolor[rgb]{0.93, 0.80, 0.80} 17.7 &\cellcolor[rgb]{0.99, 0.99, 1.00} 34.6 &\cellcolor[rgb]{0.99, 0.99, 1.00} 34.9 &\cellcolor[rgb]{0.96, 0.97, 1.00} 37.5 &\cellcolor[rgb]{0.94, 0.96, 0.99} 40.4 &\cellcolor[rgb]{0.81, 0.87, 0.98} 53.8 &\cellcolor[rgb]{0.99, 0.97, 0.97} 30.9 &\cellcolor[rgb]{0.89, 0.68, 0.68} 8.9 &\cellcolor[rgb]{0.94, 0.82, 0.82} 19.6 &\cellcolor[rgb]{0.94, 0.82, 0.82} 19.6 &\cellcolor[rgb]{0.95, 0.85, 0.85} 21.9 &\cellcolor[rgb]{0.96, 0.87, 0.87} 23.6 &\cellcolor[rgb]{0.99, 1.00, 1.00} 33.9 &\cellcolor[rgb]{0.92, 0.78, 0.78} 16.2 & 33.1 \\
Gemini-1.5-flash &\cellcolor[rgb]{0.85, 0.90, 0.98} 49.7 &\cellcolor[rgb]{0.77, 0.84, 0.97} 58.1 &\cellcolor[rgb]{0.77, 0.84, 0.97} 58.9 &\cellcolor[rgb]{0.74, 0.82, 0.97} 61.8 &\cellcolor[rgb]{0.73, 0.82, 0.97} 62.6 &\cellcolor[rgb]{0.71, 0.80, 0.97} 65.1 &\cellcolor[rgb]{0.76, 0.84, 0.97} 59.4 &\cellcolor[rgb]{0.93, 0.79, 0.79} 17.4 &\cellcolor[rgb]{0.99, 0.98, 0.98} 31.9 &\cellcolor[rgb]{0.99, 0.98, 0.98} 31.8 &\cellcolor[rgb]{0.99, 0.99, 1.00} 34.2 &\cellcolor[rgb]{0.91, 0.94, 0.99} 43.6 &\cellcolor[rgb]{0.83, 0.88, 0.98} 51.7 &\cellcolor[rgb]{1.00, 0.99, 0.99} 32.8 &\cellcolor[rgb]{0.89, 0.69, 0.69} 9.2 &\cellcolor[rgb]{0.94, 0.82, 0.82} 19.4 &\cellcolor[rgb]{0.94, 0.83, 0.83} 20.0 &\cellcolor[rgb]{0.95, 0.85, 0.85} 22.0 &\cellcolor[rgb]{0.98, 0.94, 0.94} 28.7 &\cellcolor[rgb]{0.99, 0.99, 1.00} 34.9 &\cellcolor[rgb]{0.94, 0.84, 0.84} 21.0 & 31.6 \\
Command-r + &\cellcolor[rgb]{0.90, 0.93, 0.99} 44.2 &\cellcolor[rgb]{0.79, 0.85, 0.98} 56.4 &\cellcolor[rgb]{0.82, 0.88, 0.98} 53.1 &\cellcolor[rgb]{0.79, 0.86, 0.98} 56.2 &\cellcolor[rgb]{0.77, 0.84, 0.97} 58.9 &\cellcolor[rgb]{0.75, 0.83, 0.97} 61.0 &\cellcolor[rgb]{0.90, 0.93, 0.99} 44.5 &\cellcolor[rgb]{0.94, 0.83, 0.83} 20.4 &\cellcolor[rgb]{0.92, 0.95, 0.99} 41.7 &\cellcolor[rgb]{0.92, 0.95, 0.99} 41.7 &\cellcolor[rgb]{0.91, 0.94, 0.99} 43.1 &\cellcolor[rgb]{0.88, 0.92, 0.99} 46.8 &\cellcolor[rgb]{0.75, 0.83, 0.97} 60.2 &\cellcolor[rgb]{0.94, 0.82, 0.82} 19.9 &\cellcolor[rgb]{0.89, 0.69, 0.69} 9.6 &\cellcolor[rgb]{0.96, 0.89, 0.89} 24.7 &\cellcolor[rgb]{0.96, 0.88, 0.88} 24.0 &\cellcolor[rgb]{0.97, 0.90, 0.90} 25.7 &\cellcolor[rgb]{0.98, 0.95, 0.95} 29.3 &\cellcolor[rgb]{0.95, 0.97, 0.99} 38.3 &\cellcolor[rgb]{0.89, 0.69, 0.69} 9.7 & 25.5 \\
Claude 3 Sonnet &\cellcolor[rgb]{0.79, 0.86, 0.98} 55.8 &\cellcolor[rgb]{0.67, 0.76, 0.94} 70.6 &\cellcolor[rgb]{0.68, 0.76, 0.95} 69.7 &\cellcolor[rgb]{0.66, 0.74, 0.93} 72.1 &\cellcolor[rgb]{0.66, 0.73, 0.93} 73.1 &\cellcolor[rgb]{0.63, 0.69, 0.90} 77.7 &\cellcolor[rgb]{0.65, 0.73, 0.93} 73.6 &\cellcolor[rgb]{0.93, 0.80, 0.80} 18.0 &\cellcolor[rgb]{0.99, 0.99, 1.00} 34.9 &\cellcolor[rgb]{0.97, 0.98, 1.00} 36.6 &\cellcolor[rgb]{0.96, 0.98, 1.00} 37.3 &\cellcolor[rgb]{0.93, 0.95, 0.99} 41.1 &\cellcolor[rgb]{0.83, 0.88, 0.98} 51.7 &\cellcolor[rgb]{0.96, 0.87, 0.87} 23.5 &\cellcolor[rgb]{0.90, 0.71, 0.71} 11.0 &\cellcolor[rgb]{0.97, 0.91, 0.91} 26.1 &\cellcolor[rgb]{0.97, 0.92, 0.92} 27.2 &\cellcolor[rgb]{0.98, 0.94, 0.94} 28.5 &\cellcolor[rgb]{0.99, 0.97, 0.97} 31.4 &\cellcolor[rgb]{0.93, 0.95, 0.99} 41.2 &\cellcolor[rgb]{0.93, 0.80, 0.80} 18.3 & 33.5 \\
Claude 3 Opus &\cellcolor[rgb]{0.79, 0.85, 0.98} 56.5 &\cellcolor[rgb]{0.66, 0.74, 0.93} 72.4 &\cellcolor[rgb]{0.68, 0.76, 0.95} 69.6 &\cellcolor[rgb]{0.66, 0.74, 0.93} 72.5 &\cellcolor[rgb]{0.64, 0.70, 0.91} 76.5 &\cellcolor[rgb]{0.61, 0.66, 0.88} 81.4 &\cellcolor[rgb]{0.64, 0.71, 0.91} 76.2 &\cellcolor[rgb]{0.93, 0.80, 0.80} 17.7 &\cellcolor[rgb]{0.99, 0.99, 1.00} 34.3 &\cellcolor[rgb]{0.98, 0.98, 1.00} 35.8 &\cellcolor[rgb]{0.96, 0.98, 1.00} 37.3 &\cellcolor[rgb]{0.94, 0.96, 0.99} 39.4 &\cellcolor[rgb]{0.84, 0.89, 0.98} 50.7 &\cellcolor[rgb]{0.95, 0.86, 0.86} 22.3 &\cellcolor[rgb]{0.90, 0.71, 0.71} 11.1 &\cellcolor[rgb]{0.97, 0.91, 0.91} 26.5 &\cellcolor[rgb]{0.97, 0.91, 0.91} 26.7 &\cellcolor[rgb]{0.98, 0.94, 0.94} 28.6 &\cellcolor[rgb]{0.99, 0.98, 0.98} 31.9 &\cellcolor[rgb]{0.92, 0.94, 0.99} 42.5 &\cellcolor[rgb]{0.93, 0.80, 0.80} 18.0 & 29.3 \\
GPT-4o &\cellcolor[rgb]{0.81, 0.87, 0.98} 54.0 &\cellcolor[rgb]{0.69, 0.79, 0.96} 67.1 &\cellcolor[rgb]{0.69, 0.78, 0.96} 67.8 &\cellcolor[rgb]{0.70, 0.79, 0.97} 66.6 &\cellcolor[rgb]{0.67, 0.76, 0.94} 70.4 &\cellcolor[rgb]{0.64, 0.70, 0.91} 76.6 &\cellcolor[rgb]{0.70, 0.79, 0.97} 66.1 &\cellcolor[rgb]{0.95, 0.85, 0.85} 21.9 &\cellcolor[rgb]{0.95, 0.97, 0.99} 38.4 &\cellcolor[rgb]{0.96, 0.97, 1.00} 38.0 &\cellcolor[rgb]{0.95, 0.97, 0.99} 38.6 &\cellcolor[rgb]{0.93, 0.95, 0.99} 41.3 &\cellcolor[rgb]{0.81, 0.87, 0.98} 54.6 &\cellcolor[rgb]{0.92, 0.78, 0.78} 16.2 &\cellcolor[rgb]{0.91, 0.73, 0.73} 12.6 &\cellcolor[rgb]{0.97, 0.92, 0.92} 27.3 &\cellcolor[rgb]{0.97, 0.93, 0.93} 27.6 &\cellcolor[rgb]{0.97, 0.92, 0.92} 27.3 &\cellcolor[rgb]{0.99, 0.97, 0.97} 30.8 &\cellcolor[rgb]{0.91, 0.94, 0.99} 43.4 &\cellcolor[rgb]{0.90, 0.71, 0.71} 11.4 & 36.5 \\
Gemini-1.5-pro &\cellcolor[rgb]{0.82, 0.88, 0.98} 53.0 &\cellcolor[rgb]{0.72, 0.81, 0.97} 63.5 &\cellcolor[rgb]{0.71, 0.80, 0.97} 64.9 &\cellcolor[rgb]{0.72, 0.81, 0.97} 63.6 &\cellcolor[rgb]{0.68, 0.77, 0.96} 68.4 &\cellcolor[rgb]{0.69, 0.78, 0.96} 67.6 &\cellcolor[rgb]{0.68, 0.76, 0.95} 70.0 &\cellcolor[rgb]{0.95, 0.85, 0.85} 21.9 &\cellcolor[rgb]{0.91, 0.94, 0.99} 43.1 &\cellcolor[rgb]{0.90, 0.93, 0.99} 44.5 &\cellcolor[rgb]{0.88, 0.92, 0.99} 46.6 &\cellcolor[rgb]{0.85, 0.90, 0.98} 49.7 &\cellcolor[rgb]{0.72, 0.81, 0.97} 64.1 &\cellcolor[rgb]{0.84, 0.89, 0.98} 51.0 &\cellcolor[rgb]{0.91, 0.73, 0.73} 12.3 &\cellcolor[rgb]{0.98, 0.94, 0.94} 28.6 &\cellcolor[rgb]{0.99, 0.97, 0.97} 31.0 &\cellcolor[rgb]{0.99, 0.97, 0.97} 30.8 &\cellcolor[rgb]{0.98, 0.98, 1.00} 36.0 &\cellcolor[rgb]{0.90, 0.93, 0.99} 44.6 &\cellcolor[rgb]{0.96, 0.97, 1.00} 37.8 & 30.2 \\

    \cmidrule(r{15pt}){1-1} \cmidrule(r{10pt}){2-8} \cmidrule(r{10pt}){9-15} \cmidrule(){16-22} \cmidrule(lr){23-23}
    
    Human Perf. & -- & -- & -- & -- & -- & \cellcolor[rgb]{0.66, 0.74, 0.93} 74.5 & -- & -- & -- & -- & -- & -- & \cellcolor[rgb]{0.64, 0.70, 0.91} 73.9 & -- & -- & -- & -- & -- & -- & \cellcolor[rgb]{0.80, 0.86, 0.98} 56.1 & -- & 29.7 \\

    \bottomrule
    \end{tabular}
    }
    
    \caption{Summary of a Haystack results of human performance, RAG systems, and Long-Context LLMs. Results are reported using three metrics: Coverage (left), Citation (center), and Joint (right) scores. \texttt{Full} corresponds to model performance when inputting the entire Haystack, whereas Rand, Vect, LongE, KWs, RR3, Orac correspond to retrieval components RAG systems. Models ranked by Oracle Joint Score. For each model, \#$W_{b}$ report the average number of words per bullet point.}
    \vspace{-0.1in}
    \label{tab:main_results}
\end{table*}

%% file: 5-experiments.tex
\section{Results} \label{sec:results}

\subsection{Experimental Settings} \label{sec:exp_settings}

As illustrated in Figure~\ref{fig:haystack_synthesis} (right), we evaluate both long-context LLMs that directly access the full Haystack and RAG systems where a retriever filters the Haystack down to documents it perceives as relevant, which get passed to a generator (LLM). By default, documents in the Haystack are ordered in a single arbitrary order.

\par \textbf{Full-Context Summarization}
We test a range of recent LLMs with context lengths longer than an individual Haystack, including Cohere's Command-R and Command-R+, Google's Gemini-1.5-pro and Gemini-1.5-flash \cite{reid2024gemini}, OpenAI's GPT4-turbo and GPT-4o, and Anthropic's Claude3 models (haiku, sonnet, and opus).
We also include GPT-3.5 exclusively in the RAG setting, as its context length is 16k tokens.

\par \textbf{Retrieval-Augmented Summarization}
We evaluate RAG systems to reduce the Haystack input size. 
All retrieval models receive the query and all Haystack documents and must produce a query relevance score for each document.
We sort the documents in reverse order according to the query relevance score and select the first 15k worth of document tokens. 
We chose 15k enables us to experiment with generators that have a 16k context window (GPT-3.5-turbo). We experiment with a total of six retrievers, each implemented as a separate query relevance score function. Under \textbf{KWs}, the document score is the number of overlapping keywords, extracted by NLTK, between the document and the subtopic query.
We compare embedding methods that compute the cosine similarity between each document and the subtopic query, \textbf{Vect}, a SentenceTransformers \cite{reimers-2019-sentence-bert} embedder and \textbf{LongE} \cite{zhu2024longembed}, which extends standard embedders to cover longer input contexts, and include the Rerank3 (\textbf{RR3}) model from Cohere \cite{cohere2024rerank}.
We also include a \textbf{Rand} baseline that randomly assigns relevance scores, and an oracle setting ranker \textbf{Orac}, whose score is the number of subtopic insights that appear in a given document.
The last two provide lower- and upper-bound retrieval quality estimates.

\begin{figure*}
    \centering
    \includegraphics[width=0.95\textwidth]{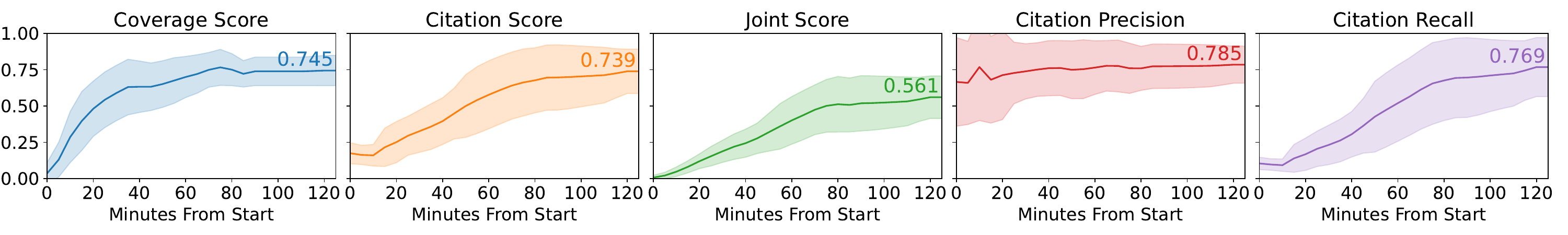}
    \caption{Estimates of human performance on the SummHay task, plotted over time as participants complete the task in the Oracle setting during two-hour sessions.}
    \label{fig:human_perf}
\end{figure*}

\subsection{Benchmark Results}

Table~\ref{tab:main_results} summarizes the SummHay results across long-context, in the \texttt{Full} column, and RAG settings for all 10 Summarizers included in our study.

Coverage scores -- which measure the presence of expected insights in a system's output summary -- range from 36.2\% when using a random retriever and GPT3.5-turbo as the summarizer, to 81.4\% using the Oracle retriever with the Claude 3 Opus summarizer.
The choice of retriever impacts the Coverage score, with Random and Oracle retrievers leading to the best and worst scores, respectively, for almost all summarizers.
Yet, top-performing LLMs like Claude3-opus achieve strong Coverage scores (70\%+) under most retrieval settings, including Full context.
In other words, \textbf{achieving strong coverage of key insights in a large corpus of text does not require retrieval, given a sufficiently capable long-context LLM}.

Citation scores -- which account for both the precision and the thoroughness of the model's attribution back to source documents -- present a complementary narrative.
The lowest citation score often occurs in the full-context setting, with citation quality on par with random retrieval. On the other hand, as retrieval improves (for example from Random to RR3 to Oracle), citation scores increase.
In a nutshell, \textbf{for use-cases where citation quality is important, optimizing retrieval is paramount}: it removes irrelevant documents from the summarizer's context, narrowing and focusing options for citation.
Gemini-1.5-pro stands out as an outlier, as it is the only model that achieves comparable Citation scores in RAG and long-context settings.

Taking Coverage and Citation into account, the Joint Score provides the complete system performance on SummHay.
As expected, all Summarizers perform best with the Oracle retriever, an unrealistic setting intended to evaluate score ranges.

All models except for Gemini-1.5-pro achieve their realistic best performance in the RAG setting using the RR3 retriever.
The higher relative performance of the more advanced RAG retriever RR3, developed for enterprise search and RAG settings, aligns with expectations compared to simpler retrievers. 
This result confirms the validity of our SummHay as a test-bed for holistic RAG evaluation; newer, more advanced RAGs can be benchmarked in an end-to-end fashion on SummHay, measuring direct impact on output quality.

We still observe two large gaps: all RAG systems underperform the Oracle setting, indicating ample room for improvements in RAG systems, and \textbf{models achieve very low Joint scores in the full-context setting (10-20), indicating that SummHay is an unsolved task for long-context LLMs.} Gemini-1.5-pro stands with strong ability to cite in the full-context setting, achieving the only realistic score above 35 on the benchmark.

In the Rerank3 RAG setting, the top three models are neck and neck, with Claude3 Opus, GPT-4o, and Gemini-1.5-pro all achieving Joint Scores between 30.8 and 36.0. Yet these models achieve these performances through different trade-offs, with Claude3 Opus obtaining the highest Coverage, Gemini-1.5-pro the highest Citation, and GPT-4o at the mid-point. This confirms there is room to grow: \textbf{a system that achieves the coverage of Claude3 Opus with the citation quality of Gemini-1.5-pro can exceed current score by 15-20\%}. 

The right-most column of Table~\ref{tab:main_results} shows each system's average bullet point length (number of tokens). Several systems (Gemini-1.5-pro and Claude 3 Opus) average 30 words per bullet, close human-written bullet points (29.7). Others (GPT-4o, GPT4-turbo) are more verbose, at 36-38 words per bullet point. In Appendix~\ref{app:auto_eval_bias} we confirm that verbosity does not bias evaluation: succinct methods can achieve strong performance on SummHay.

Appendix~\ref{app:summhay_prec_rec} breaks down Citation Scores into Precision and Recall. No system excels at either precision or recall but we do observe trade-offs. For example, Claude models generally achieve higher precision and lower recall, whereas Command-r + and GPT-4o favor recall over precision.

\subsection{Estimating Human Performance}
\label{subsec:human_perf}

We estimate human performance on the task by recruiting two annotators to perform the task. We first define the setting in which annotation was conducted, then go over the results.

The participants performed the task in the Oracle setting, only viewing documents relevant to the query they are currently summarizing, as it reduces the volume of text they must read from about 100,000 tokens to 15,000 tokens. We assume this effectively reduces the amount of time required for annotation by a factor of 5-6, but this remains unverified, as it is impractical to conduct human annotation on the full Haystack.

In total, two annotators participated in writing a total of 10 summaries, five for subtopics in the conversational domain, and five for subtopics in the news domain. Although this represents a subset of the 92 subtopics in the entire SummHay benchmark, we believe it represents an unbiased estimate of human performance on the benchmark.

Figure~\ref{fig:human_perf} aggregates results across the ten annotation sessions. Overall, participants make steady progress during the sessions, with both Coverage and Citation rising rapidly in the first 90 minutes, and then at a slower pace in the last 30 minutes.

The Citation Score corresponds to an F1 measure, and we also report on the Precision and Recalls of the Citations. We find that citation precision averages close to 80.0 throughout the session, whereas recall rises steadily during the session.

Table~\ref{tab:main_results} includes the summarized final scores in contrast to system scores, showing that a human annotator can significantly outperform LLMs and RAG systems on the SummHay task, as the human joint score (56.1) is significantly higher than the best system performance (44.6). We caution the reader not to consider our estimate of human performance as an upper bound, as we believe that with more time and explicit instructions to double-check their work, annotators could further increase their scores. We solely intend the human performance to be a reference point for achievable performances on the benchmark, and we expect future systems to tie and surpass human performance on the task. Appendix~\ref{app:summhay_human_perf} provides further detail on guidelines, recruitment, and task framing.

\input{tables/sort_sensitivity}

\subsection{Position Bias Sensitivity}
In the Full Context experiment results (Table~\ref{tab:main_results}, \texttt{Full} columns), documents in the Haystack are ordered arbitrarily, with relevant documents in the top, middle, and bottom portions of the context window.
Prior work \cite{huang2023embrace,chang2023booookscore,chen2023understanding,ravaut2023context} has reported that models exhibit a \texttt{position bias} that leads them to put more importance on information in the context window's extremities.
SummHay offers a framework to study position bias systematically.
In Table~\ref{tab:sort_sentitivity}, we report the results of the Position Bias experiment, in which we run the SummHay experiment with the top-3 performing models on sorted Haystacks, where relevant documents to a subtopic are either all at the \texttt{Top} or \texttt{Bottom} of the context window.
Similar to prior work, we find that all three models exhibit position bias, with GPT-4o and Claude3 Opus performing better when relevant documents are at the bottom of the context window, and Gemini-1.5-pro favoring \texttt{Top} Haystacks. We compute a \texttt{Position Sensitivity} score as the maximum absolute difference in Joint Score between the \texttt{Random} ordering and Top and Bottom conditions. Future systems should strive to attain minimal sensitivity on SummHay, as document ordering is often arbitrary in real-world applications. 

%% file: tables/sort_sensitivity.tex
\begin{table}[]
    \centering
    \resizebox{0.40\textwidth}{!}{%
    \begin{tabular}{lcccc}
    & \multicolumn{3}{c}{\textbf{Document Order}} & \\
    \cmidrule(lr){2-4}
    \textbf{Summarizer} & Top & Bottom & Random & \textbf{Sensitivity} \\
    \midrule
    GPT-4o & 13.8 & \cellcolor[rgb]{0.80, 0.86, 0.98} \textbf{24.1} & 11.4 & 12.6 \\
    Claude3 Opus & 20.4 & \cellcolor[rgb]{0.80, 0.86, 0.98} \textbf{28.0} & 18.0 & 10.0 \\
    Gemini-1.5-pro & \cellcolor[rgb]{0.80, 0.86, 0.98} \textbf{47.1} & 38.9 & 37.9 & \textbf{9.2} \\
    \bottomrule
    \end{tabular}
    }
    \caption{Joint Scores of LLMs in the Full Context Setting, based on how documents are sorted. Documents can be in \texttt{Random} order or sorted such that relevant ones are at the \texttt{Top} or \texttt{Bottom} of the context window.}
    \vspace{-0.2in}
    \label{tab:sort_sentitivity}
\end{table}

%% file: 6-ablations.tex

%% file: 7-discussion.tex
\section{Discussion} \label{sec:discussion}

\paragraph{Task Upper Bound}
In Section \ref{subsec:haystack_gen} and Appendix \ref{app:haystack_verification}, we detail our data pipeline and efforts to ensure the quality of our dataset. 
This includes insight subtopic verification and verifying the inclusion of only specified insights for each document. 
Despite our efforts to prevent overlap among insights and guarantee the presence of insights in the documents generated, errors may be introduced from using LLMs in scaling this data synthesis, making achieving a perfect joint score of 100 likely unachievable.
Although we estimate human performance in a simplified setting, we do not determine the task upper bound. 
However, we show significant room for improvement between the realistic full-context and RAG settings and the Oracle setting.

\paragraph{Simplifying Assumptions in Data Synthesis}
The assumptions made when generating Haystack documents likely introduce artificial signals that simplify the task. 
For example, in order to maximize control over the data synthesis process, Haystack documents are generated independently; no dependencies or cross-references exist among the documents. 
However, in a real-world multi-document summarization task, documents may link or refer to each other, and there may be temporal between documents.
We believe the introduction of more realistic assumptions can further increase the difficulty of the task, and we hope that future work will take our synthesis processes as a starting point for such improvements.


\paragraph{Controlling for Verbosity}
When generating summaries, we specify the desired number of insights for the LLM to generate.
Furthermore, we do not control for or penalize the verbosity of the summaries, and summaries with longer insights may result in higher coverage.
Not specifying the number of summary insights needed per query will result in a more difficult task, and we leave a study of the potential trade-offs between verbosity, human preference, and overall scores for future work.

\paragraph{Reliance on Automated Evaluation}
Although we do not observe significant bias towards a particular family of models, as shown in Appendix \ref{app:auto_eval_bias}, the results in Table \ref{tab:auto_eval_results} demonstrate that there is room for improvement both in coverage and linking evaluation. 
Gemini-1.5-pro, in addition to being more costly than GPT-4o, had a rate limit which inhibited its use in our study. 
Although highly-cost effective, non-LLM based NLI and relevance metrics \cite{chen2023menli,liu2023towards} were not tested; \citet{chen2023understanding} found worse performance among smaller NLI models for the related task of unrelated sentence identification on long-form question answering. 
%
%
\paragraph{Model Choice}
The generation models included in our study are all closed-source models. 
Although these closed-source models currently generally outperform open-sourced models, this performance comparison can be task-dependent \cite{chen2023chatgpt}. 
We exclude high-performing open-sourced models such as Llama-3 \cite{llama3modelcard} as the original models cannot handle the minimal 16k context window necessary for our RAG experiments.
Restricting the length of the retrieved documents to, for example, 8k would remove too many of the insights for a given subtopic; by allowing up to 15k tokens in the RAG setting, we find that the oracle citation F1 achievable with this context length is 0.84 averaged across insights, which we believe strikes a balance between the reduction of input size and the feasibility of the task.
We leave an analysis of RAG systems across retrieved input lengths, as well as models specifically designed for output citation \cite{menick2022teaching} for future work and encourage and benchmarking of longer-context, open-sourced models on SummHay.


%% file: 8-conclusion.tex
\section{Conclusion} \label{sec:conclusion}
In this work, we address the challenges of evaluating long-context LLMs and RAG systems by introducing the SummHay benchmark task, synthesized to assess the ability of systems to precisely summarize large sets of documents.
The SummHay task requires generating summaries that accurately cover and cite insights relevant to a particular query.
Our comprehensive evaluation reveals that current models struggle with this task; even in an oracle document setting, models lag behind human performance by more than 10 points.
We believe that SummHay provides a robust framework for evaluating long-context systems and will encourage researchers to utilize SummHay to drive progress toward systems that can match or surpass human performance in long-context summarization.

%% file: 9-limitations.tex

\section*{Ethical Considerations}

The models and datasets utilized in the project primarily reflect the culture of the English-speaking populace. Gender, age, race, and other socio-economic biases may exist in the data, and models trained on these datasets may propagate these biases. Text generation tasks such as summarization have previously been shown to contain these biases.

In Section~\ref{sec:evaluation} and Section~\ref{sec:results}, we recruited professional annotators to perform evaluation, or directly attempt the task. We ensured to remunerate the participants fairly (\$25/hour). Participants could communicate with us to voice concerns, work at their own pace, and choose to stop working on the project at any time. Finally, we ensured to anonymize the annotations (annotator identity is instead marked as \texttt{annotator1}, \texttt{annotator2}, etc.).

In our work, we relied on several datasets as well as pre-trained language models. We explicitly verified that all datasets and models are publicly released for research purposes and that we have proper permission to reuse and modify the datasets.

%% file: 10-appendix.tex
\section{Appendix}
\label{sec:appendix}

\subsection{Haystack Synthesis Details}
\label{app:haystack_verification}
Below we specify additional details regarding our data synthesis pipelines. 
For the news domain, we leverage GPT-4o for all data synthesis steps.
We found it necessary to leverage this high-performing LLM due to the longer seed context documents that subtopics and insights are generated from. For the conversation domain, we leverage GPT-4o to generate subtopics and insights, and for any LLM-based verification step. Conversation generation (conditioned on selected insights) is completed using GPT-3.5-turbo.

Furthermore, we employ verification steps to ensure that subtopics and insights are distinct and precisely mapped to Haystack documents.
When generating documents given a set of insights, we do not want other insights to ``\textit{leak}'' into the document, as that would reduce the quality of the Haystack and task.
Below we list the verification steps taken across the conversation and news domain. 
Differences in domain characteristics and the seed used to generate Haystacks necessitate per-domain verification steps.
%
%

\subsubsection{Subtopic Verification}
\label{app:subtopic_verification}
In the news domain, to ensure distinct insights and subtopics we first prompt an LLM to identify any overlapping or duplicate subtopics and remove these subtopics. 
This helps ensure that when querying relevant insights for a subtopic, insights can only belong to one of the distinct subtopics generated.

In the conversation domain, we use manual inspection to verify the distinctness of the subtopics and regenerate subtopic candidates (at temperature $T=1$) until obtaining a list where each subtopic feels qualitatively unique.

\subsubsection{Insight Verification}
\label{app:insight_verification}
For the news domain, we prompt the LLM to remove duplicate insights.
After producing an initial set of insights for the subtopics, we take all insights and prompt an LLM to categorize each insight into one of the subtopics. 
As insights are initially generated for a particular subtopic, at this step we ensure that no insight can fit into another subtopic. 
We thus remove any insights for which the categorized subtopic differs from the one it was initially generated for.

In the conversation domain, we iterate over subtopics sequentially and use a prompt to generate the list of insights for one subtopic. In the prompt, we provide not only the target subtopic but also all other subtopics and insights, instructing the LLM to avoid such subtopics and insights, and only propose insights that are distinct and unique in contrast to those. Manual inspection from the authors reveals that: as long as the subtopics are confirmed to be unique, and that the insights are enforced to be specific (and include entities), very little overlap occurs across subtopic insights.
\subsubsection{Document Verification}
\label{app:document_verificdation}

In the news domain, to ensure a precise mapping of insights to documents in which an insight is present, we prompt an LLM to label whether any of the insights, across subtopics, other than those sampled for that document are found in the document. 
If any are found, we regenerate the document, asking the LLM to remove the sentence(s) containing the extraneous insights.
This procedure is repeated until no extraneous insights are found, up until 5 iterations. 
For a similar purpose, we prompt an LLM to label whether the insights sampled for a given document are indeed found in the document. 
If not, we regenerate the document to add sentences that contain this insight, up until 5 iterations. 
We find that after 5 iterations of editing, discrepancies in insights by the LLM were primarily paraphrasing or partial detail upon manual inspection.

In the conversational domain, we generate the documents iteratively, one \textit{chapter} at a time, where each chapter is intended to introduce a single insight. When expanding an insight into a chapter, we generate a candidate chapter and use GPT-4o to classify whether the candidate chapter indeed covers the expected subtopic and insight. If not, we regenerate the candidate chapter up to ten times, and otherwise, we accept the candidate chapter into the document and proceed with the next chapter. We find that in practice, the generation process requires 5 iterations less than 1\% of the time to generate a chapter that GPT-4o can correctly assign to the expected insight. 

\subsection{Evaluation Prompt} \label{app:eval_prompt}

Below, we list the prompt we use for automatic evaluation in the SummHay task, as described in Section~\ref{sec:evaluation}.

\begin{lstlisting}
You are given a list of bullet points (each with a unique number), and a specific reference insight. Your objective is to determine whether the reference insight is covered in any of the bullet points. You must further determine if the insight is partially covered ("PARTIAL_COVERAGE") or fully covered ("FULL_COVERAGE") by the bullet points. If the insight is not covered at all, you must return "NO_COVERAGE". See examples below:

[[FEW_SHOT_EXAMPLES]]

Now complete the task for the following insight and bullet points:

Reference Insight:
[[INSIGHT]]

Bullet Points:
[[BULLETS]]

Requirements:
- Do not hallucinate that the insight is covered by the bullet points if it is not.
- Your response should only be the JSON output in the format above, such that it can directly parsed by Python's json module. DO NOT OUTPUT ANY EXPLANATION OR ANYTHING THAT IS NOT THE JSON RESPONSE.

\end{lstlisting}

\subsection{Automatic Results Bias} \label{app:auto_eval_bias}

\input{tables/auto_eval_bias}

There is a concern that since we propose to use an LLM to automate the evaluation of the SummHay experiment, the choice of the evaluator model might affect the validity of the results if such a model has a systematic bias in its judgment. We evaluate the possible presence of two biases. First, whether the automatic evaluation could favor outputs of one model family over the other (e.g., GPT-4o systematically favoring outputs of the GPT* family).

To study this, we perform an automatic evaluation of score bias by calculating the difference $(\Delta)$ between the Coverage Score according to the auto-evaluation, and the Coverage Score from the manual annotation. A positive $\Delta$ indicates that the auto-evaluation assigned a higher score than the manual evaluation, and a negative $\Delta$ indicates the opposite. We calculate the average $\Delta$ for each Summarizer model in our experiment, and inspect the bias of three evaluator models: GPT-4o, Claude3 Opus, and Gemini-1.5-pro.

The top portion of Table~\ref{tab:auto_eval_bias} summarizes the Summarizer model bias analysis results. First, we find that auto-evaluation results almost always have a positive bias, indicating that on average, the auto-evaluation overestimates Coverage by roughly 5 points across models. Second, we find that evaluator models tend to have a positive bias for top-performing Summarizer models (e.g., GPT-4o, Claude3 Opus, and Gemini-1.5-pro), but do not systematically prefer outputs from a specific model family. In fact, Claude3 Opus seems to be particularly critical of Claude3 model outputs (with biases very close to zero). All models have a strong positive bias towards GPT-4o outputs, but it does not translate to bias for a model family. \textbf{Overall, we find no evidence of systematic bias of automatic evaluation that would favor one model family over the other.} The analysis does reveal a pattern of overestimating coverage by an average of 5 points, which should be taken into account when interpreting the results.

In a second analysis, we study whether automatic evaluation leads to favoring summaries based on their length.

To study length bias, we first see whether a summary's score correlates with its length, as measured by the number of words divided by the number of bullet points. We divide by the number of bullet points as each query requires a different number of bullet points, which directly affects length of the summary in a way that is not controlled by the LLM. By measuring length bias based on the length of individual bullet points, we remove this confounding variable.

In the \texttt{Length to Score Corr.} row of Table~\ref{tab:auto_eval_bias}, we find that there is a slight negative correlation (-0.12 to -0.178) between a summary's score and the number of words in its bullet points. This correlation could be explained by results from Section~\ref{sec:results}, which showed that several of the top-performing models (Claude3-opus, Gemini-1.5-pro) generate the shortest bullet points. In other words, short bullet points could achieve higher scores not because they are short, but because they were generated by better models. To remove this confounding variable, we measure whether automatic score $\Delta$ (computed above) correlates with bullet point length. This analysis, summarized in te \texttt{Length to Delta Corr.} row of Table~\ref{tab:auto_eval_bias}, indicates a non-existent correlation between bullet-point length, and whether the automatic evaluator was biased in its scoring (-0.081 to 0.02). In conclusion, we do not find evidence that using automatic evaluation with our evaluation protocol will cause length bias in our results, which would systems to generate shorter or longer bullets points.

\subsection{Details on Establishing SummHay Human Performance} \label{app:summhay_human_perf}


To establish task duration, we considered an average reading speed of 200 words per minute. Carefully reading the documents (roughly 12,000 words) would therefore require one hour. Accounting for the need to write the summary, and scan multiple times over documents to identify and cite insights, annotators were told they had a maximum of two hours to complete the task. Participants could take breaks (i.e., pause their work) and finish the task early if they felt the task was completed. After their initial sessions, participants were asked whether two hours seemed appropriate to complete the task without rushing, and both agreed. Participants sometimes used the entirety of the two hours, and in other cases completed the task in as little as 80 minutes.

We relied on professional annotators known and trusted by our research group (based on performance on previous annotation work), and they were compensated at 25 USD per hour. One of the annotators participated in the annotation of automated summaries (and therefore had access to reference insights for certain subtopics), and we ensured that this annotator only performed the summarization task for subtopics and document sets they had not seen during that annotation, ensuring they had no prior knowledge of the documents in the Haystack.

Participants were given all documents in a shuffled order and were instructed to read them carefully and summarize any insight that seemed to be repeating across documents.
Participants were told the number of present insights (similar to LLM prompt in our experiment).
In practice, we found that participants chose to write slightly more bullet points than the number they were given.

Regarding citation, participants were instructed to be as thorough as possible and to explicitly look for additional citations when they had identified an insight.

Regarding tool use, participants were allowed to use string search (i.e., \texttt{Ctrl+F}), but were prohibited from copying the text from the documents, and were explicitly instructed they should not use ChatGPT or equivalent LLM-based interfaces in any way to assist them with the task. Because this cannot be strictly enforced practically, we rely on trusted professional annotators to complete the task.

Participants were instructed to gradually write their summary in a text box in the annotation interface, and we recorded the progress on the summary during each annotation summary. We then performed auto-evaluation using the same settings used in our benchmarking experiments on the final summary of the session, as well as a summary every 10 minutes during the session.

\subsection{Citation Precision \& Recall Analysis} \label{app:summhay_prec_rec}

\input{tables/precision_recall}

The Citation Score in the SummHay benchmark is an F1 calculation between the set of cites generated by a system in a given bullet point, and the expected cites of the matched reference insight, based on knowledge from the Haystack generation of what documents include the insight. F1 is chosen as a measure to ensure that systems balance between precise and thorough cites.

Table~\ref{tab:prec_rec} reports the precision and recall of all systems on the benchmark, as well as the F1 (i.e., the Citation score) to shed light on how different systems balance between precision and recall.

\subsection{Model Access Details}
For each model in our study, we specify its model card and how it was accessed. 

We access the Google models Gemini-1.5-pro (\texttt{gemini-1.5-pro-preview-0514}) and Gemini-1.5-flash (\texttt{gemini-1.5-flash-preview-0514}) through Vertex AI \footnote{\url{https://cloud.google.com/vertex-ai}}.

We include three OpenAI models in our study: GPT-3.5-turbo (\texttt{gpt-3.5-turbo-0125}), GPT-4-turbo (\texttt{gpt-4-turbo-2024-04-09}), and GPT-4o (\texttt{gpt-4o}).
All models were accessed through OpenAI's official API\footnote{\url{https://github.com/openai/opeai-python}}.

Cohere summarizers Command-R (\texttt{cohere.command-r-v1:0}) and Command-R+ (\texttt{cohere.command-r-plus-v1:0}) were accessed through Amazon Bedrock\footnote{\url{https://aws.amazon.com/bedrock/}}, while Rerank3 (\texttt{rerank-english-v3.0}) was accessed through Cohere's official API\footnote{\url{https://docs.cohere.com/reference/rerank}}.

Anthropic models were also accessed through Amazon Bedrock: Claude 3 Haiku (\texttt{anthropic.claude-3-haiku-20240307-v1:0}), Claude 3 Sonnet (\texttt{anthropic.claude-3-sonnet-20240229-v1:0}), and Claude 3 Opus (\texttt{anthropic.claude-3-opus-20240229-v1:0}). 

The embedders Vect (\texttt{sentence-transformers/all-mpnet-base-v2"}) and LongEmbed (\texttt{dwzhu/e5-base-4k}) were accessed through SentenceTransformers \cite{reimers-2019-sentence-bert} and huggingface's transformers library \cite{wolf2019huggingface}, respectively.

\subsection{Additional Output Examples} \label{app:additional_examples}

\begin{figure*}
    \centering
    \includegraphics[width=0.99\textwidth]{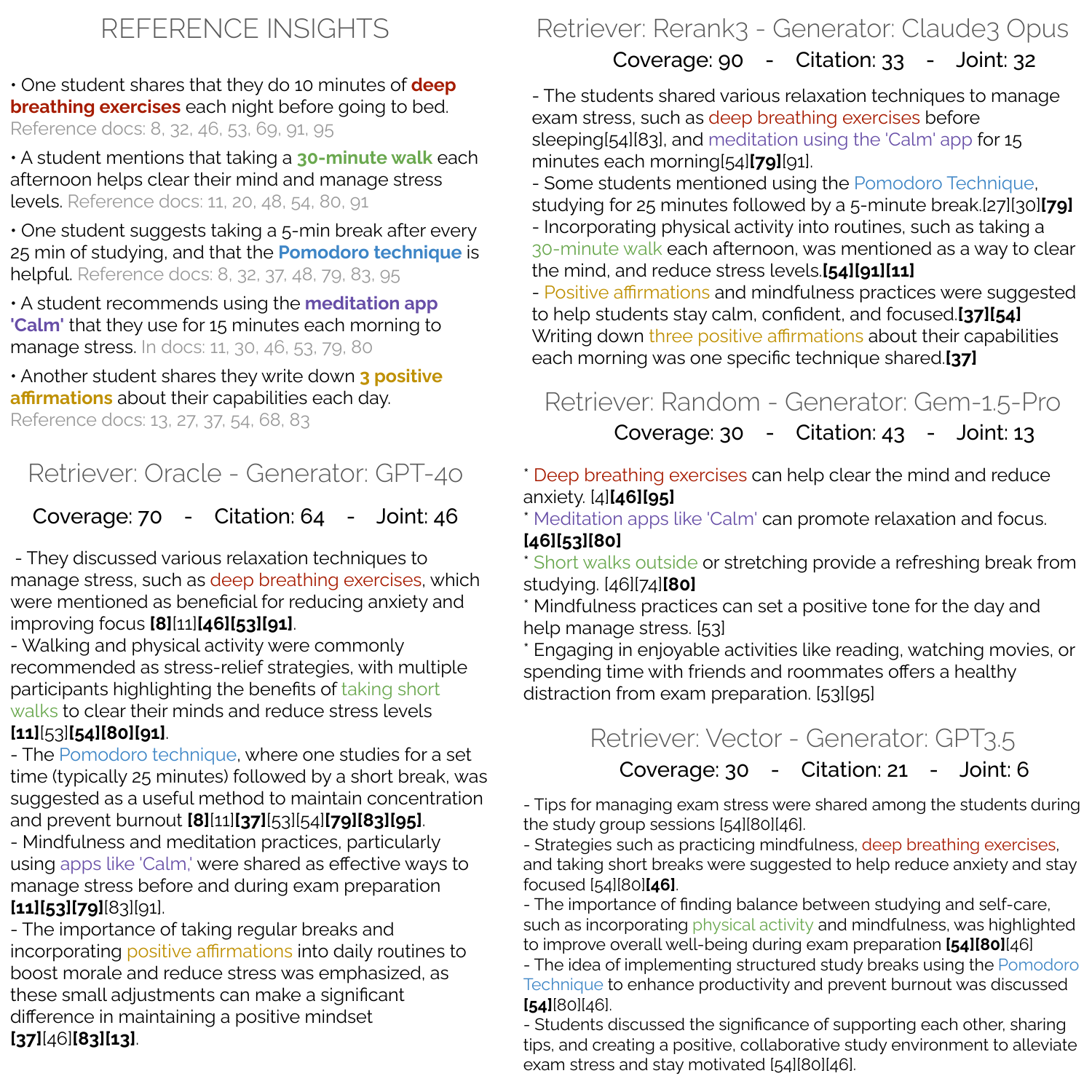}
    \caption{Examples of five insights within a subtopic, and four SummHay outputs from RAG systems, including the final Coverage, Citation, and Joint scores, as calculated by our LLM-based automatic evaluation. We add color coding and bolding to facilitate the interpretation of the evaluation.}
    \label{fig:additional_examples}
\end{figure*}

Figure~\ref{fig:additional_examples} provides four examples of real summary outputs from different RAG pipelines on a common subtopic related to managing stress when preparing to an exam. For each summary, we also report on the Coverage, Citation and Joint scores, as calculated by the LLM-based automatic evaluation. We add color coding and bolding to facilitate the interpretation of the evaluation.
\subsection{Additional Discussion}
We point to several additional areas for future work.
\paragraph{English-centric}
While our data pipeline can be extended to non-English languages with access to a seed scenario in a given language, our benchmark was developed only on English and may be more reliable in English.
However, the task is language-agnostic, and future work can create a multilingual version of the SummHay task, similar to efforts such as Seahorse  \cite{clark2023seahorse}.
\paragraph{Beyond Relevance}
We restrict our data synthesis process and analysis to target summarization relevance, but we believe that similar data procedures and evaluations could be applied to factual consistency. We leave an extension of our pipeline and analysis of model outputs along other dimensions such as coherence, efficiency (brevity), or factuality to future work.

\paragraph{Focus on Factoid-Style Insights.} Our Haystack synthesis process incentivizes the creation of specific insights that focus on a number or entity. Specificity helps simplify evaluation, and ensure we can achieve reproducible automatic evaluation. Yet real-world scenarios might have less clear-cut insights, with different documents only partially overlapping on insights, or with potentially disagreeing conclusions (e.g., some people like the Pomodoro Technique while others don't). Prior work has shown that NLP methods struggle in such cases of coverage diversity \cite{laban2022discord,huang2023embrace}, and including such discord within the Haystack could yield more complex and realistic tasks.

\paragraph{Investigating Low Scores.} Manual inspection reveals different failure modes in low-scoring summaries, including (1) retrieving insights that are not relevant to the query (other subtopics), (2) framing very high-level information as a specific insight (e.g., ``all the participants are polite to each other''), (3) hallucinating insights not directly supported in the documents. We do not systematically evaluate the frequency of each failure, but future work can explore this more systematically, for example using efficient NLI-based alignment \cite{laban2022summac,zha2023alignscore,tang2024minicheck}.

%% file: tables/auto_eval_bias.tex
\begin{table}[]
    \centering
    \resizebox{0.48\textwidth}{!}{%
    \begin{tabular}{lccc}
    & \multicolumn{3}{c}{\textbf{Evaluator Model}} \\
    \cmidrule(lr){2-4}
     & GPT-4o & Opus & Gem-1.5-Pro \\
    \midrule
    \multicolumn{4}{c}{\textbf{Summarizer Model Bias}} \\
    \midrule

Claude 3 Sonnet &\cellcolor[rgb]{0.74, 0.82, 0.97} 0.027 &\cellcolor[rgb]{0.67, 0.76, 0.94} -0.001 &\cellcolor[rgb]{0.67, 0.76, 0.94} -0.012 \\
Gemini-1.5-flash &\cellcolor[rgb]{0.85, 0.90, 0.98} 0.051 &\cellcolor[rgb]{0.73, 0.81, 0.97} 0.024 &\cellcolor[rgb]{0.67, 0.76, 0.94} -0.009 \\
GPT3.5 &\cellcolor[rgb]{0.67, 0.76, 0.94} 0.009 &\cellcolor[rgb]{0.85, 0.90, 0.98} 0.050 &\cellcolor[rgb]{0.84, 0.89, 0.98} 0.048 \\
Claude 3 Opus &\cellcolor[rgb]{0.89, 0.92, 0.99} 0.059 &\cellcolor[rgb]{0.77, 0.84, 0.97} 0.034 &\cellcolor[rgb]{0.82, 0.87, 0.98} 0.043 \\
Gemini-1.5-pro &\cellcolor[rgb]{0.99, 0.97, 0.97} 0.088 &\cellcolor[rgb]{0.92, 0.94, 0.99} 0.065 &\cellcolor[rgb]{0.92, 0.94, 0.99} 0.065 \\
GPT4-turbo &\cellcolor[rgb]{0.96, 0.97, 1.00} 0.075 &\cellcolor[rgb]{0.98, 0.95, 0.95} 0.091 &\cellcolor[rgb]{0.87, 0.91, 0.99} 0.056 \\
Command-r + &\cellcolor[rgb]{0.91, 0.94, 0.99} 0.064 &\cellcolor[rgb]{0.90, 0.71, 0.71} 0.128 &\cellcolor[rgb]{0.94, 0.96, 0.99} 0.071 \\
Claude 3 Haiku &\cellcolor[rgb]{0.98, 0.94, 0.94} 0.092 &\cellcolor[rgb]{0.94, 0.84, 0.84} 0.108 &\cellcolor[rgb]{0.94, 0.96, 0.99} 0.071 \\
GPT-4o &\cellcolor[rgb]{0.97, 0.91, 0.91} 0.097 &\cellcolor[rgb]{0.96, 0.88, 0.88} 0.102 &\cellcolor[rgb]{0.95, 0.85, 0.85} 0.106 \\
\hline
Average &\cellcolor[rgb]{0.90, 0.93, 0.99} 0.062 &\cellcolor[rgb]{0.93, 0.95, 0.99} 0.067 &\cellcolor[rgb]{0.84, 0.89, 0.98} 0.049 \\

    \midrule
    \multicolumn{4}{c}{\textbf{Summary Length Bias}} \\
    \midrule
    Length to Score Corr. & \cellcolor[rgb]{0.91, 0.94, 0.99} -0.122 & 
 \cellcolor[rgb]{0.96, 0.97, 1.00} -0.174 & \cellcolor[rgb]{0.96, 0.97, 1.00} -0.178 \\
    Length to Delta Corr. & \cellcolor[rgb]{0.67, 0.76, 0.94} 0.02 & \cellcolor[rgb]{0.74, 0.82, 0.97} -0.051 & \cellcolor[rgb]{0.84, 0.89, 0.98} -0.081 \\
    \bottomrule
    \end{tabular}
    }
    \caption{Results of analysis of potential bias in automated evaluation. We explore potential bias caused by what model is used, which is reported in differences of scores with human annotation, and bias due to the length of the summary, which is reported as a correlation. An unbiased evaluator model should achieve a bias close to zero on both analyses.}
    \label{tab:auto_eval_bias}
\end{table}

%% file: tables/precision_recall.tex
\begin{table}[]
    \centering
    \resizebox{0.49\textwidth}{!}{%
    \begin{tabular}{lcccccc}
     & \multicolumn{2}{c}{\textbf{Precision}} & \multicolumn{2}{c}{\textbf{Recall}} & \multicolumn{2}{c}{\textbf{F1 (Citation)}} \\
    \cmidrule(lr){2-3} \cmidrule(lr){4-5} \cmidrule(lr){6-7}
    \textbf{Summarizer} & Orac & Full & Orac & Full & Orac & Full \\
    \cmidrule(lr){1-1} \cmidrule(lr){2-3} \cmidrule(lr){4-5} \cmidrule(lr){6-7}
    GPT3.5 &\cellcolor[rgb]{0.88, 0.92, 0.99} 46.7 & -- &\cellcolor[rgb]{0.93, 0.80, 0.80} 17.9 & -- &\cellcolor[rgb]{0.95, 0.87, 0.87} 23.0 & -- \\
    Claude 3 Haiku &\cellcolor[rgb]{0.85, 0.90, 0.98} 49.3 &\cellcolor[rgb]{0.96, 0.89, 0.89} 24.7 &\cellcolor[rgb]{0.99, 0.98, 0.98} 31.6 &\cellcolor[rgb]{0.96, 0.88, 0.88} 24.2 &\cellcolor[rgb]{0.98, 0.99, 1.00} 35.6 &\cellcolor[rgb]{0.91, 0.75, 0.75} 14.1 \\
    GPT4-turbo &\cellcolor[rgb]{0.73, 0.82, 0.97} 62.3 &\cellcolor[rgb]{0.91, 0.75, 0.75} 14.1 &\cellcolor[rgb]{0.98, 0.99, 1.00} 35.7 &\cellcolor[rgb]{0.87, 0.61, 0.61} 3.8 &\cellcolor[rgb]{0.93, 0.95, 0.99} 41.4 &\cellcolor[rgb]{0.87, 0.64, 0.64} 5.5 \\
    Claude 3 Opus &\cellcolor[rgb]{0.70, 0.79, 0.97} 66.0 &\cellcolor[rgb]{0.99, 0.97, 0.97} 30.7 &\cellcolor[rgb]{0.89, 0.92, 0.99} 45.8 &\cellcolor[rgb]{0.96, 0.88, 0.88} 24.0 &\cellcolor[rgb]{0.84, 0.89, 0.98} 50.7 &\cellcolor[rgb]{0.95, 0.86, 0.86} 22.3 \\
    Gemini-1.5-flash &\cellcolor[rgb]{0.78, 0.85, 0.97} 57.8 &\cellcolor[rgb]{0.96, 0.97, 0.99} 38.2 &\cellcolor[rgb]{0.81, 0.87, 0.98} 54.5 &\cellcolor[rgb]{0.90, 0.93, 0.99} 44.2 &\cellcolor[rgb]{0.83, 0.88, 0.98} 51.7 &\cellcolor[rgb]{1.00, 0.99, 0.99} 32.8 \\
    Claude 3 Sonnet &\cellcolor[rgb]{0.69, 0.78, 0.96} 67.3 &\cellcolor[rgb]{0.92, 0.94, 0.99} 42.2 &\cellcolor[rgb]{0.87, 0.91, 0.99} 47.2 &\cellcolor[rgb]{0.94, 0.82, 0.82} 19.9 &\cellcolor[rgb]{0.83, 0.88, 0.98} 51.7 &\cellcolor[rgb]{0.96, 0.87, 0.87} 23.5 \\
    Command-r &\cellcolor[rgb]{0.77, 0.84, 0.97} 58.9 &\cellcolor[rgb]{0.95, 0.96, 0.99} 38.9 &\cellcolor[rgb]{0.79, 0.86, 0.98} 55.9 &\cellcolor[rgb]{0.99, 0.98, 0.98} 31.7 &\cellcolor[rgb]{0.81, 0.87, 0.98} 53.8 &\cellcolor[rgb]{0.99, 0.97, 0.97} 30.9 \\
    GPT-4o &\cellcolor[rgb]{0.70, 0.80, 0.97} 65.7 &\cellcolor[rgb]{0.98, 0.93, 0.93} 28.0 &\cellcolor[rgb]{0.84, 0.89, 0.98} 51.0 &\cellcolor[rgb]{0.91, 0.73, 0.73} 13.0 &\cellcolor[rgb]{0.81, 0.87, 0.98} 54.6 &\cellcolor[rgb]{0.92, 0.78, 0.78} 16.2 \\
    Command-r + &\cellcolor[rgb]{0.69, 0.78, 0.96} 67.6 &\cellcolor[rgb]{0.96, 0.88, 0.88} 24.2 &\cellcolor[rgb]{0.75, 0.83, 0.97} 60.8 &\cellcolor[rgb]{0.95, 0.84, 0.84} 21.2 &\cellcolor[rgb]{0.75, 0.83, 0.97} 60.2 &\cellcolor[rgb]{0.94, 0.82, 0.82} 19.9 \\
    Gemini-1.5-pro &\cellcolor[rgb]{0.64, 0.71, 0.91} 76.2 &\cellcolor[rgb]{0.77, 0.84, 0.97} 59.0 &\cellcolor[rgb]{0.75, 0.83, 0.97} 60.9 &\cellcolor[rgb]{0.83, 0.88, 0.98} 52.4 &\cellcolor[rgb]{0.72, 0.81, 0.97} 64.1 &\cellcolor[rgb]{0.84, 0.89, 0.98} 51.0 \\
    \hline
    Human Perf. & \cellcolor[rgb]{0.61, 0.67, 0.89} 78.8 & -- & \cellcolor[rgb]{0.61, 0.67, 0.89} 82.4 & -- & \cellcolor[rgb]{0.64, 0.70, 0.91} 76.7 & -- \\
    \bottomrule
    \end{tabular}
    }

    \caption{Breakdown of Citation Precision and Recall of models on the SummHay Benchmark, combined into an F1 Score (Citation Score). Numbers are reported for the Full Context and the Oracle settings of SummHay.}
    \label{tab:prec_rec}
\end{table}

%% file: acl_latex.bbl
\begin{thebibliography}{59}
\providecommand{\natexlab}[1]{#1}

\bibitem[{AI@Meta(2024)}]{llama3modelcard}
AI@Meta. 2024.
\newblock \href {https://github.com/meta-llama/llama3/blob/main/MODEL_CARD.md} {Llama 3 model card}.

\bibitem[{An et~al.(2023)An, Gong, Zhong, Li, Zhang, Kong, and Qiu}]{an2023eval}
Chenxin An, Shansan Gong, Ming Zhong, Mukai Li, Jun Zhang, Lingpeng Kong, and Xipeng Qiu. 2023.
\newblock L-eval: Instituting standardized evaluation for long context language models.
\newblock \emph{arXiv preprint arXiv:2307.11088}.

\bibitem[{Bai et~al.(2023)Bai, Lv, Zhang, Lyu, Tang, Huang, Du, Liu, Zeng, Hou et~al.}]{bai2023longbench}
Yushi Bai, Xin Lv, Jiajie Zhang, Hongchang Lyu, Jiankai Tang, Zhidian Huang, Zhengxiao Du, Xiao Liu, Aohan Zeng, Lei Hou, et~al. 2023.
\newblock Longbench: A bilingual, multitask benchmark for long context understanding.
\newblock \emph{arXiv preprint arXiv:2308.14508}.

\bibitem[{Beltagy et~al.(2020)Beltagy, Peters, and Cohan}]{beltagy2020longformer}
Iz~Beltagy, Matthew~E Peters, and Arman Cohan. 2020.
\newblock Longformer: The long-document transformer.
\newblock \emph{arXiv preprint arXiv:2004.05150}.

\bibitem[{Bhandari et~al.(2020)Bhandari, Narayan~Gour, Ashfaq, Liu, and Neubig}]{Bhandari-2020-reevaluating}
Manik Bhandari, Pranav Narayan~Gour, Atabak Ashfaq, Pengfei Liu, and Graham Neubig. 2020.
\newblock Re-evaluating evaluation in text summarization.
\newblock In \emph{Proceedings of the 2020 Conference on Empirical Methods in Natural Language Processing (EMNLP)}.

\bibitem[{Chang et~al.(2023)Chang, Lo, Goyal, and Iyyer}]{chang2023booookscore}
Yapei Chang, Kyle Lo, Tanya Goyal, and Mohit Iyyer. 2023.
\newblock Booookscore: A systematic exploration of book-length summarization in the era of llms.
\newblock \emph{arXiv preprint arXiv:2310.00785}.

\bibitem[{Chen et~al.(2023{\natexlab{a}})Chen, Jiao, Li, Qin, Ravaut, Zhao, Xiong, and Joty}]{chen2023chatgpt}
Hailin Chen, Fangkai Jiao, Xingxuan Li, Chengwei Qin, Mathieu Ravaut, Ruochen Zhao, Caiming Xiong, and Shafiq Joty. 2023{\natexlab{a}}.
\newblock Chatgpt's one-year anniversary: Are open-source large language models catching up?
\newblock \emph{arXiv preprint arXiv:2311.16989}.

\bibitem[{Chen et~al.(2023{\natexlab{b}})Chen, Xu, Arora, and Choi}]{chen2023understanding}
Hung-Ting Chen, Fangyuan Xu, Shane~A Arora, and Eunsol Choi. 2023{\natexlab{b}}.
\newblock Understanding retrieval augmentation for long-form question answering.
\newblock \emph{arXiv preprint arXiv:2310.12150}.

\bibitem[{Chen and Eger(2023)}]{chen2023menli}
Yanran Chen and Steffen Eger. 2023.
\newblock Menli: Robust evaluation metrics from natural language inference.
\newblock \emph{Transactions of the Association for Computational Linguistics}, 11:804--825.

\bibitem[{Clark et~al.(2023)Clark, Rijhwani, Gehrmann, Maynez, Aharoni, Nikolaev, Sellam, Siddhant, Das, and Parikh}]{clark2023seahorse}
Elizabeth Clark, Shruti Rijhwani, Sebastian Gehrmann, Joshua Maynez, Roee Aharoni, Vitaly Nikolaev, Thibault Sellam, Aditya Siddhant, Dipanjan Das, and Ankur~P Parikh. 2023.
\newblock Seahorse: A multilingual, multifaceted dataset for summarization evaluation.
\newblock \emph{arXiv preprint arXiv:2305.13194}.

\bibitem[{Dong et~al.(2023)Dong, Tang, Li, Zhao, and Wen}]{dong2023bamboo}
Zican Dong, Tianyi Tang, Junyi Li, Wayne~Xin Zhao, and Ji-Rong Wen. 2023.
\newblock Bamboo: A comprehensive benchmark for evaluating long text modeling capacities of large language models.
\newblock \emph{arXiv preprint arXiv:2309.13345}.

\bibitem[{Fabbri et~al.(2021{\natexlab{a}})Fabbri, Kry{\'s}ci{\'n}ski, McCann, Xiong, Socher, and Radev}]{fabbri2021summeval}
Alexander~R Fabbri, Wojciech Kry{\'s}ci{\'n}ski, Bryan McCann, Caiming Xiong, Richard Socher, and Dragomir Radev. 2021{\natexlab{a}}.
\newblock Summeval: Re-evaluating summarization evaluation.
\newblock \emph{Transactions of the Association for Computational Linguistics}, 9:391--409.

\bibitem[{Fabbri et~al.(2021{\natexlab{b}})Fabbri, Kry{\'s}ci{\'n}ski, McCann, Xiong, Socher, and Radev}]{fabbri-etal-2021-summeval}
Alexander~R. Fabbri, Wojciech Kry{\'s}ci{\'n}ski, Bryan McCann, Caiming Xiong, Richard Socher, and Dragomir Radev. 2021{\natexlab{b}}.
\newblock \href {https://doi.org/10.1162/tacl_a_00373} {{S}umm{E}val: Re-evaluating summarization evaluation}.
\newblock \emph{Transactions of the Association for Computational Linguistics}, 9:391--409.

\bibitem[{Fan et~al.(2019)Fan, Jernite, Perez, Grangier, Weston, and Auli}]{fan-etal-2019-eli5}
Angela Fan, Yacine Jernite, Ethan Perez, David Grangier, Jason Weston, and Michael Auli. 2019.
\newblock \href {https://doi.org/10.18653/v1/P19-1346} {{ELI}5: Long form question answering}.
\newblock In \emph{Proceedings of the 57th Annual Meeting of the Association for Computational Linguistics}, pages 3558--3567, Florence, Italy. Association for Computational Linguistics.

\bibitem[{Gao and Wan(2022)}]{gao-wan-2022-dialsummeval}
Mingqi Gao and Xiaojun Wan. 2022.
\newblock \href {https://doi.org/10.18653/v1/2022.naacl-main.418} {{D}ial{S}umm{E}val: Revisiting summarization evaluation for dialogues}.
\newblock In \emph{Proceedings of the 2022 Conference of the North American Chapter of the Association for Computational Linguistics: Human Language Technologies}, pages 5693--5709, Seattle, United States. Association for Computational Linguistics.

\bibitem[{Gliwa et~al.(2019)Gliwa, Mochol, Biesek, and Wawer}]{gliwa-etal-2019-samsum}
Bogdan Gliwa, Iwona Mochol, Maciej Biesek, and Aleksander Wawer. 2019.
\newblock \href {https://doi.org/10.18653/v1/D19-5409} {{SAMS}um corpus: A human-annotated dialogue dataset for abstractive summarization}.
\newblock In \emph{Proceedings of the 2nd Workshop on New Frontiers in Summarization}, pages 70--79, Hong Kong, China. Association for Computational Linguistics.

\bibitem[{Guu et~al.(2020)Guu, Lee, Tung, Pasupat, and Chang}]{guu2020retrieval}
Kelvin Guu, Kenton Lee, Zora Tung, Panupong Pasupat, and Mingwei Chang. 2020.
\newblock Retrieval augmented language model pre-training.
\newblock In \emph{International conference on machine learning}, pages 3929--3938. PMLR.

\bibitem[{Hermann et~al.(2015)Hermann, Kociský, Grefenstette, Espeholt, Kay, Suleyman, and Blunsom}]{DBLP:conf/nips/HermannKGEKSB15}
Karl~Moritz Hermann, Tomás Kociský, Edward Grefenstette, Lasse Espeholt, Will Kay, Mustafa Suleyman, and Phil Blunsom. 2015.
\newblock \href {http://papers.nips.cc/paper/5945-teaching-machines-to-read-and-comprehend} {Teaching machines to read and comprehend}.
\newblock In \emph{NIPS}, pages 1693--1701.

\bibitem[{Hu et~al.(2024)Hu, Chen, Wu, Qi, Bi, Wu, and Pan}]{hu2024benchmarking}
Nan Hu, Jiaoyan Chen, Yike Wu, Guilin Qi, Sheng Bi, Tongtong Wu, and Jeff~Z Pan. 2024.
\newblock Benchmarking large language models in complex question answering attribution using knowledge graphs.
\newblock \emph{arXiv preprint arXiv:2401.14640}.

\bibitem[{Huang et~al.(2023)Huang, Laban, Fabbri, Choubey, Joty, Xiong, and Wu}]{huang2023embrace}
Kung-Hsiang Huang, Philippe Laban, Alexander~R Fabbri, Prafulla~Kumar Choubey, Shafiq Joty, Caiming Xiong, and Chien-Sheng Wu. 2023.
\newblock Embrace divergence for richer insights: A multi-document summarization benchmark and a case study on summarizing diverse information from news articles.
\newblock \emph{arXiv preprint arXiv:2309.09369}.

\bibitem[{Inc.(2024)}]{cohere2024rerank}
Cohere Inc. 2024.
\newblock \href {https://cohere.com/blog/rerank-3} {Introducing rerank 3: The next generation of search relevance}.
\newblock Accessed: 2024-06-10.

\bibitem[{Kamalloo et~al.(2023)Kamalloo, Jafari, Zhang, Thakur, and Lin}]{kamalloo2023hagrid}
Ehsan Kamalloo, Aref Jafari, Xinyu Zhang, Nandan Thakur, and Jimmy Lin. 2023.
\newblock Hagrid: A human-llm collaborative dataset for generative information-seeking with attribution.
\newblock \emph{arXiv preprint arXiv:2307.16883}.

\bibitem[{Kamradt(2023)}]{kamradt2023}
Gregory Kamradt. 2023.
\newblock \href {https://github.com/gkamradt/LLMTest_NeedleInAHaystack/blob/main/README.md} {Needleinahaystack}.

\bibitem[{Kim et~al.(2024)Kim, Chang, Karpinska, Garimella, Manjunatha, Lo, Goyal, and Iyyer}]{kim2024fables}
Yekyung Kim, Yapei Chang, Marzena Karpinska, Aparna Garimella, Varun Manjunatha, Kyle Lo, Tanya Goyal, and Mohit Iyyer. 2024.
\newblock Fables: Evaluating faithfulness and content selection in book-length summarization.
\newblock \emph{arXiv preprint arXiv:2404.01261}.

\bibitem[{Krishna et~al.(2023)Krishna, Bransom, Kuehl, Iyyer, Dasigi, Cohan, and Lo}]{krishna2023longeval}
Kalpesh Krishna, Erin Bransom, Bailey Kuehl, Mohit Iyyer, Pradeep Dasigi, Arman Cohan, and Kyle Lo. 2023.
\newblock Longeval: Guidelines for human evaluation of faithfulness in long-form summarization.
\newblock \emph{arXiv preprint arXiv:2301.13298}.

\bibitem[{Kuratov et~al.(2024)Kuratov, Bulatov, Anokhin, Sorokin, Sorokin, and Burtsev}]{kuratov2024search}
Yuri Kuratov, Aydar Bulatov, Petr Anokhin, Dmitry Sorokin, Artyom Sorokin, and Mikhail Burtsev. 2024.
\newblock In search of needles in a 10m haystack: Recurrent memory finds what llms miss.
\newblock \emph{arXiv preprint arXiv:2402.10790}.

\bibitem[{Kwan et~al.(2023)Kwan, Zeng, Wang, Sun, Li, Shang, Liu, and Wong}]{kwan2023m4le}
Wai-Chung Kwan, Xingshan Zeng, Yufei Wang, Yusen Sun, Liangyou Li, Lifeng Shang, Qun Liu, and Kam-Fai Wong. 2023.
\newblock M4le: A multi-ability multi-range multi-task multi-domain long-context evaluation benchmark for large language models.
\newblock \emph{arXiv preprint arXiv:2310.19240}.

\bibitem[{Laban et~al.(2020)Laban, Hsi, Canny, and Hearst}]{laban2020summary}
Philippe Laban, Andrew Hsi, John Canny, and Marti~A Hearst. 2020.
\newblock The summary loop: Learning to write abstractive summaries without examples.
\newblock In \emph{Proceedings of the 58th Annual Meeting of the Association for Computational Linguistics}, pages 5135--5150.

\bibitem[{Laban et~al.(2022{\natexlab{a}})Laban, Schnabel, Bennett, and Hearst}]{laban2022summac}
Philippe Laban, Tobias Schnabel, Paul~N Bennett, and Marti~A Hearst. 2022{\natexlab{a}}.
\newblock Summac: Re-visiting nli-based models for inconsistency detection in summarization.
\newblock \emph{Transactions of the Association for Computational Linguistics}, 10:163--177.

\bibitem[{Laban et~al.(2022{\natexlab{b}})Laban, Wu, Murakhovs’ka, Chen, and Xiong}]{laban2022discord}
Philippe Laban, Chien-Sheng Wu, Lidiya Murakhovs’ka, Xiang Chen, and Caiming Xiong. 2022{\natexlab{b}}.
\newblock Discord questions: A computational approach to diversity analysis in news coverage.
\newblock In \emph{Findings of the Association for Computational Linguistics: EMNLP 2022}, pages 5180--5194.

\bibitem[{LangChain(2024)}]{multineedle}
LangChain. 2024.
\newblock \href {https://blog.langchain.dev/multi-needle-in-a-haystack/?ref=dailydev} {Multi-needle in a haystack}.

\bibitem[{Lewis et~al.(2019)Lewis, Liu, Goyal, Ghazvininejad, Mohamed, Levy, Stoyanov, and Zettlemoyer}]{lewis2019bart}
Mike Lewis, Yinhan Liu, Naman Goyal, Marjan Ghazvininejad, Abdelrahman Mohamed, Omer Levy, Ves Stoyanov, and Luke Zettlemoyer. 2019.
\newblock Bart: Denoising sequence-to-sequence pre-training for natural language generation, translation, and comprehension.
\newblock \emph{arXiv preprint arXiv:1910.13461}.

\bibitem[{Lewis et~al.(2020)Lewis, Perez, Piktus, Petroni, Karpukhin, Goyal, K{\"u}ttler, Lewis, Yih, Rockt{\"a}schel et~al.}]{lewis2020retrieval}
Patrick Lewis, Ethan Perez, Aleksandra Piktus, Fabio Petroni, Vladimir Karpukhin, Naman Goyal, Heinrich K{\"u}ttler, Mike Lewis, Wen-tau Yih, Tim Rockt{\"a}schel, et~al. 2020.
\newblock Retrieval-augmented generation for knowledge-intensive nlp tasks.
\newblock \emph{Advances in Neural Information Processing Systems}, 33:9459--9474.

\bibitem[{Li et~al.(2023{\natexlab{a}})Li, Sun, Hu, Liu, Chen, Hu, Wu, and Zhang}]{li2023survey}
Dongfang Li, Zetian Sun, Xinshuo Hu, Zhenyu Liu, Ziyang Chen, Baotian Hu, Aiguo Wu, and Min Zhang. 2023{\natexlab{a}}.
\newblock A survey of large language models attribution.
\newblock \emph{arXiv preprint arXiv:2311.03731}.

\bibitem[{Li et~al.(2023{\natexlab{b}})Li, Cao, Pan, Ma, and Sun}]{li2023towards}
Xinze Li, Yixin Cao, Liangming Pan, Yubo Ma, and Aixin Sun. 2023{\natexlab{b}}.
\newblock Towards verifiable generation: A benchmark for knowledge-aware language model attribution.
\newblock \emph{arXiv preprint arXiv:2310.05634}.

\bibitem[{Li et~al.(2024)Li, Yue, Liao, and Sun}]{li2024attributionbench}
Yifei Li, Xiang Yue, Zeyi Liao, and Huan Sun. 2024.
\newblock Attributionbench: How hard is automatic attribution evaluation?
\newblock \emph{arXiv preprint arXiv:2402.15089}.

\bibitem[{Liu et~al.(2022)Liu, Fabbri, Liu, Zhao, Nan, Han, Han, Joty, Wu, Xiong et~al.}]{liu2022revisiting}
Yixin Liu, Alexander~R Fabbri, Pengfei Liu, Yilun Zhao, Linyong Nan, Ruilin Han, Simeng Han, Shafiq Joty, Chien-Sheng Wu, Caiming Xiong, et~al. 2022.
\newblock Revisiting the gold standard: Grounding summarization evaluation with robust human evaluation.
\newblock \emph{arXiv preprint arXiv:2212.07981}.

\bibitem[{Liu et~al.(2023)Liu, Fabbri, Zhao, Liu, Joty, Wu, Xiong, and Radev}]{liu2023towards}
Yixin Liu, Alexander~R Fabbri, Yilun Zhao, Pengfei Liu, Shafiq Joty, Chien-Sheng Wu, Caiming Xiong, and Dragomir Radev. 2023.
\newblock Towards interpretable and efficient automatic reference-based summarization evaluation.
\newblock \emph{arXiv preprint arXiv:2303.03608}.

\bibitem[{Machlab and Battle(2024)}]{machlab2024llm}
Daniel Machlab and Rick Battle. 2024.
\newblock Llm in-context recall is prompt dependent.
\newblock \emph{arXiv preprint arXiv:2404.08865}.

\bibitem[{Menick et~al.(2022)Menick, Trebacz, Mikulik, Aslanides, Song, Chadwick, Glaese, Young, Campbell-Gillingham, Irving et~al.}]{menick2022teaching}
Jacob Menick, Maja Trebacz, Vladimir Mikulik, John Aslanides, Francis Song, Martin Chadwick, Mia Glaese, Susannah Young, Lucy Campbell-Gillingham, Geoffrey Irving, et~al. 2022.
\newblock Teaching language models to support answers with verified quotes.
\newblock \emph{arXiv preprint arXiv:2203.11147}.

\bibitem[{Min et~al.(2023)Min, Krishna, Lyu, Lewis, Yih, Koh, Iyyer, Zettlemoyer, and Hajishirzi}]{min2023factscore}
Sewon Min, Kalpesh Krishna, Xinxi Lyu, Mike Lewis, Wen-tau Yih, Pang~Wei Koh, Mohit Iyyer, Luke Zettlemoyer, and Hannaneh Hajishirzi. 2023.
\newblock Factscore: Fine-grained atomic evaluation of factual precision in long form text generation.
\newblock \emph{arXiv preprint arXiv:2305.14251}.

\bibitem[{Ni et~al.(2024)Ni, Cai, Wei, Wang, Yin, and Li}]{ni2024xl}
Xuanfan Ni, Hengyi Cai, Xiaochi Wei, Shuaiqiang Wang, Dawei Yin, and Piji Li. 2024.
\newblock Xl bench: A benchmark for extremely long context understanding with long-range dependencies.
\newblock \emph{arXiv preprint arXiv:2404.05446}.

\bibitem[{Raffel et~al.(2020)Raffel, Shazeer, Roberts, Lee, Narang, Matena, Zhou, Li, and Liu}]{raffel2020exploring}
Colin Raffel, Noam Shazeer, Adam Roberts, Katherine Lee, Sharan Narang, Michael Matena, Yanqi Zhou, Wei Li, and Peter~J Liu. 2020.
\newblock Exploring the limits of transfer learning with a unified text-to-text transformer.
\newblock \emph{Journal of machine learning research}, 21(140):1--67.

\bibitem[{Ravaut et~al.(2023)Ravaut, Joty, Sun, and Chen}]{ravaut2023context}
Mathieu Ravaut, Shafiq Joty, Aixin Sun, and Nancy~F Chen. 2023.
\newblock On context utilization in summarization with large language models.
\newblock \emph{arXiv e-prints}, pages arXiv--2310.

\bibitem[{Reid et~al.(2024)Reid, Savinov, Teplyashin, Lepikhin, Lillicrap, Alayrac, Soricut, Lazaridou, Firat, Schrittwieser et~al.}]{reid2024gemini}
Machel Reid, Nikolay Savinov, Denis Teplyashin, Dmitry Lepikhin, Timothy Lillicrap, Jean-baptiste Alayrac, Radu Soricut, Angeliki Lazaridou, Orhan Firat, Julian Schrittwieser, et~al. 2024.
\newblock Gemini 1.5: Unlocking multimodal understanding across millions of tokens of context.
\newblock \emph{arXiv preprint arXiv:2403.05530}.

\bibitem[{Reimers and Gurevych(2019)}]{reimers-2019-sentence-bert}
Nils Reimers and Iryna Gurevych. 2019.
\newblock \href {http://arxiv.org/abs/1908.10084} {Sentence-bert: Sentence embeddings using siamese bert-networks}.
\newblock In \emph{Proceedings of the 2019 Conference on Empirical Methods in Natural Language Processing}. Association for Computational Linguistics.

\bibitem[{Shaham et~al.(2023)Shaham, Ivgi, Efrat, Berant, and Levy}]{shaham2023zeroscrolls}
Uri Shaham, Maor Ivgi, Avia Efrat, Jonathan Berant, and Omer Levy. 2023.
\newblock Zeroscrolls: A zero-shot benchmark for long text understanding.
\newblock \emph{arXiv preprint arXiv:2305.14196}.

\bibitem[{Song et~al.(2024)Song, Chen, Chen, Yu, Wan, and Wang}]{song2024milebench}
Dingjie Song, Shunian Chen, Guiming~Hardy Chen, Fei Yu, Xiang Wan, and Benyou Wang. 2024.
\newblock Milebench: Benchmarking mllms in long context.
\newblock \emph{arXiv preprint arXiv:2404.18532}.

\bibitem[{Su et~al.(2024)Su, Ahmed, Lu, Pan, Bo, and Liu}]{su2024roformer}
Jianlin Su, Murtadha Ahmed, Yu~Lu, Shengfeng Pan, Wen Bo, and Yunfeng Liu. 2024.
\newblock Roformer: Enhanced transformer with rotary position embedding.
\newblock \emph{Neurocomputing}, 568:127063.

\bibitem[{Tang et~al.(2024)Tang, Laban, and Durrett}]{tang2024minicheck}
Liyan Tang, Philippe Laban, and Greg Durrett. 2024.
\newblock Minicheck: Efficient fact-checking of llms on grounding documents.
\newblock \emph{arXiv preprint arXiv:2404.10774}.

\bibitem[{Vig et~al.(2022)Vig, Fabbri, Kryscinski, Wu, and Liu}]{vig-etal-2022-exploring}
Jesse Vig, Alexander Fabbri, Wojciech Kryscinski, Chien-Sheng Wu, and Wenhao Liu. 2022.
\newblock \href {https://doi.org/10.18653/v1/2022.findings-naacl.109} {Exploring neural models for query-focused summarization}.
\newblock In \emph{Findings of the Association for Computational Linguistics: NAACL 2022}, pages 1455--1468, Seattle, United States. Association for Computational Linguistics.

\bibitem[{Wolf et~al.(2019)Wolf, Debut, Sanh, Chaumond, Delangue, Moi, Cistac, Rault, Louf, Funtowicz et~al.}]{wolf2019huggingface}
Thomas Wolf, Lysandre Debut, Victor Sanh, Julien Chaumond, Clement Delangue, Anthony Moi, Pierric Cistac, Tim Rault, R{\'e}mi Louf, Morgan Funtowicz, et~al. 2019.
\newblock Huggingface's transformers: State-of-the-art natural language processing.
\newblock \emph{arXiv preprint arXiv:1910.03771}.

\bibitem[{Wu et~al.(2023)Wu, Iso, Pezeshkpour, Bhutani, and Hruschka}]{wu2023less}
Yunshu Wu, Hayate Iso, Pouya Pezeshkpour, Nikita Bhutani, and Estevam Hruschka. 2023.
\newblock Less is more for long document summary evaluation by llms.
\newblock \emph{arXiv preprint arXiv:2309.07382}.

\bibitem[{Yue et~al.(2023)Yue, Wang, Chen, Zhang, Su, and Sun}]{yue2023automatic}
Xiang Yue, Boshi Wang, Ziru Chen, Kai Zhang, Yu~Su, and Huan Sun. 2023.
\newblock Automatic evaluation of attribution by large language models.
\newblock \emph{arXiv preprint arXiv:2305.06311}.

\bibitem[{Zha et~al.(2023)Zha, Yang, Li, and Hu}]{zha2023alignscore}
Yuheng Zha, Yichi Yang, Ruichen Li, and Zhiting Hu. 2023.
\newblock Alignscore: Evaluating factual consistency with a unified alignment function.
\newblock In \emph{Proceedings of the 61st Annual Meeting of the Association for Computational Linguistics (Volume 1: Long Papers)}, pages 11328--11348.

\bibitem[{Zhang et~al.(2024{\natexlab{a}})Zhang, Xu, and Perez-Beltrachini}]{zhang2024fine}
Huajian Zhang, Yumo Xu, and Laura Perez-Beltrachini. 2024{\natexlab{a}}.
\newblock Fine-grained natural language inference based faithfulness evaluation for diverse summarisation tasks.
\newblock \emph{arXiv preprint arXiv:2402.17630}.

\bibitem[{Zhang et~al.(2024{\natexlab{b}})Zhang, Chen, Hu, Xu, Chen, Hao, Han, Thai, Wang, Liu et~al.}]{zhang2024infty}
Xinrong Zhang, Yingfa Chen, Shengding Hu, Zihang Xu, Junhao Chen, Moo~Khai Hao, Xu~Han, Zhen~Leng Thai, Shuo Wang, Zhiyuan Liu, et~al. 2024{\natexlab{b}}.
\newblock Infinitybench: Extending long context evaluation beyond 100k tokens.
\newblock \emph{arXiv preprint arXiv:2402.13718}.

\bibitem[{Zhong et~al.(2021)Zhong, Yin, Yu, Zaidi, Mutuma, Jha, Awadallah, Celikyilmaz, Liu, Qiu, and Radev}]{zhong-etal-2021-qmsum}
Ming Zhong, Da~Yin, Tao Yu, Ahmad Zaidi, Mutethia Mutuma, Rahul Jha, Ahmed~Hassan Awadallah, Asli Celikyilmaz, Yang Liu, Xipeng Qiu, and Dragomir Radev. 2021.
\newblock \href {https://doi.org/10.18653/v1/2021.naacl-main.472} {{QMS}um: A new benchmark for query-based multi-domain meeting summarization}.
\newblock In \emph{Proceedings of the 2021 Conference of the North American Chapter of the Association for Computational Linguistics: Human Language Technologies}, pages 5905--5921, Online. Association for Computational Linguistics.

\bibitem[{Zhu et~al.(2024)Zhu, Wang, Yang, Song, Wu, Wei, and Li}]{zhu2024longembed}
Dawei Zhu, Liang Wang, Nan Yang, Yifan Song, Wenhao Wu, Furu Wei, and Sujian Li. 2024.
\newblock Longembed: Extending embedding models for long context retrieval.
\newblock \emph{arXiv preprint arXiv:2404.12096}.

\end{thebibliography}
